\documentclass[sigconf,manuscript,anonymous=false]{acmart}
\usepackage{graphicx}
\usepackage{subfig}

\AtBeginDocument{%
  \providecommand\BibTeX{{%
    \normalfont B\kern-0.5em{\scshape i\kern-0.25em b}\kern-0.8em\TeX}}}

\copyrightyear{2022}
\acmYear{2022}
\setcopyright{acmcopyright}\acmConference[AutomotiveUI '22]{14th International Conference on Automotive User Interfaces and Interactive Vehicular Applications}{September 17--20, 2022}{Seoul, Republic of Korea}
\acmBooktitle{14th International Conference on Automotive User Interfaces and Interactive Vehicular Applications (AutomotiveUI '22), September 17--20, 2022, Seoul, Republic of Korea}
\acmPrice{15.00}
\acmDOI{10.1145/3543174.3547117}
\acmISBN{978-1-4503-9415-4/22/09}

\setcitestyle{numbers}

\begin{document}

\title[Cause-and-Effect Analysis of ADAS]{Cause-and-Effect Analysis of ADAS: A Comparison Study between Literature Review and Complaint Data}

\author{Jackie Ayoub}
\email{jyayoub@umich.edu}
\affiliation{%
  \institution{University of Michigan Dearborn}
  \streetaddress{4901 Evergreen Road}
  \city{Dearborn}
  \state{MI}
  \postcode{48128}
  \country{USA}
}

\author{Zifei Wang}
\email{zifwang@umich.edu}
\affiliation{%
  \institution{University of Michigan Dearborn}
  \streetaddress{4901 Evergreen Road}
  \city{Dearborn}
  \state{MI}
  \postcode{48128}
  \country{USA}
}

\author{Meitang Li}
\email{meitang@umich.edu}
\affiliation{%
  \institution{University of Michigan Transportation Research Institute}
  \streetaddress{2901 Baxter Rd}
  \city{Ann Arbor}
  \state{MI}
  \postcode{48109}
  \country{USA}
}

\author{Huizhong Guo}
\email{hzhguo@umich.edu}
\affiliation{%
  \institution{University of Michigan Transportation Research Institute}
  \streetaddress{2901 Baxter Rd.}
  \city{Ann Arbor}
  \state{MI}
  \postcode{48109}
  \country{USA}
}

\author{Rini Sherony}
\email{rini.sherony@toyota.com}
\affiliation{%
 \institution{Toyota Motor North America}
 \streetaddress{1555 Woodridge Ave}
  \city{Ann Arbor}
  \state{MI}
  \postcode{48105}
  \country{USA}
  }

\author{Shan Bao}
\email{shanbao@umich.edu}
\affiliation{%
  \institution{University of Michigan Dearborn; University of Michigan Transportation Research Institute}
  \streetaddress{4901 Evergreen Road}
  \city{Dearborn}
  \state{MI}
  \postcode{48128}
  \country{USA}
}

\author{Feng Zhou}
\email{fezhou@umich.edu}
\affiliation{%
  \institution{University of Michigan Dearborn}
  \streetaddress{4901 Evergreen Road}
  \city{Dearborn}
  \state{MI}
  \postcode{48128}
  \country{USA}
}
\renewcommand{\shortauthors}{Ayoub et al.}

\begin{abstract}
Advanced driver assistance systems (ADAS) are designed to improve vehicle safety. However, it is difficult to achieve such benefits without understanding the causes and limitations of the current ADAS and their possible solutions.
This study 1) investigated the limitations and solutions of ADAS through a literature review, 2) identified the causes and effects of ADAS through consumer complaints using natural language processing models, and 3) compared the major differences between the two. 
These two lines of research identified similar categories of ADAS causes, including human factors, environmental factors, and vehicle factors. However, academic research focused more on human factors of ADAS issues and proposed advanced algorithms to mitigate such issues while drivers complained more of vehicle factors of ADAS failures, which led to associated top consequences. 
The findings from these two sources tend to complement each other and provide important implications for the improvement of ADAS in the future.  

\end{abstract}



\keywords{ADAS, cause and effect, natural language processing, literature review, complaint data}

\maketitle

\section{Introduction}

Advanced driver assistance systems (ADAS) provide warnings to drivers of an impending collision and/or temporal controls of the vehicle’s speed and/or direction under certain circumstances to assist the driver \cite{jin2019mining}. Examples of ADAS include but are not limited to Forward Collision Warning (FCW), Automatic Emergency Braking (AEB), Lane Departure Warning (LDW), Lane Keeping Assistance (LKA), Blind Spot Warning (BSW), and Adaptive Cruise Control (ACC). These ADAS are designed to take over drivers’ responsibilities and/or extend their capabilities. For example, an AEB system can brake automatically if the in-vehicle sensor detects that the vehicle is going to collide with another object in front of it and a BSW system can provide warnings if another vehicle is driving in the adjacent lane in the drivers’ blind spot. Thus, the safety benefits of ADAS, when it is implemented and used properly, are considerable because it can significantly decrease human error, economic cost, pollution \cite{winner2014handbook}. Previous studies \cite{chen2015typical, cicchino2017effectiveness, kusano2012safety, kusano2015comparison, teoh2021effectiveness} reported that vehicles with such ADAS had significantly fewer crashes and lower insurance losses compared to vehicles without ADAS.

However, the benefits of ADAS will not be reached if they are not used properly or drivers do not have proper trust in them due to possible ADAS failures \cite{ayoub2021investigation,ayoub2021modeling}. Reagan and McCartt \cite{reagan2016observed} found that under-trust and false alarms might be the main causes that led to the disuse of ADAS. For example, it was found that participants only had moderate levels of trust in LDW and LKA. Some vehicles’ FCW systems provided excessive alarms or even false alarms when the drivers did not perceive danger \cite{kidd2017driver}. Other studies found that there were negative ADAS effects on driver behavior, including distrust or overreliance on the system, wrong mental model used in the locus of control, and lack of support for behavioral adaptation, situation awareness, or mental workload \cite{richardson2001human}. For example, vehicles with ADAS might cause drivers to take a passive role in driving with less caution and attention, and thus more likely to be involved in traffic accidents \cite{stanton2005driver,zhou2020driver,zhou2020takeover}. Paul et al. \cite{pauladvanced} emphasized that the sole purpose of ADAS was to support drivers rather than to replace them. Therefore, it is imperative to examine the causes of system failures, especially those related to human factors and system designs, to identify solutions to improve ADAS design and safety. 


The overall goal of this study is to investigate ADAS failures (its causes and effects) and possible solutions through literature review and text data mining to provide recommendations for improved safety and customer experience. The overall goal of this work is divided into four specific tasks as follows:

1) Explore ADAS limitations and solutions by conducting a literature review; 

2) Build a natural language processing model based on deep learning to extract cause-and-effect relationships of customer perceptions in ADAS uses from the National Highway Traffic Safety Administration (NHTSA) consumer complaints database \cite{NHTSA2011}; 

3) Build a taxonomy of the causes and effects related to ADAS uses; 

4) Discuss the differences between the findings above to help provide recommendations for improving customer experience with ADAS.


\section{Literature Review}
We conducted a literature review to examine the limitations and proposed solutions of ADAS by searching journal articles, conference papers, and technical reports published between January 1, 2016 and June 31, 2021 using a combination of keywords in five academic databases, including Science Direct, ACM Digital Library, Scopus, IEEE Xplore, and PsycINFO. A complimentary search was performed in Google Scholar to accomplish a thorough result. The keywords included ADAS (Advanced driver/driving assistant/assistance system), FCW (forward collision warning), AEB (automatic/automated emergent/emergency brake/braking), LDW (lane departure warning/alert/prevention), LKA (lane-keeping assist/assistance), BSW (blind-spot warning/alert/monitor/assist), ACC (adaptive cruise control), automatic braking, automatic steering, automated driving, lane trace assist, supercruise, autopilot, specific names for different OEMs, lane departure prevention, and rear cross-traffic alert system with limitation, failure or improvement. In this process, we found 842 related papers initially. The following criteria were defined to select eligible papers:

1. The paper identified one or more ADAS limitations/failures;

2. The paper proposed one or more solutions for an existing ADAS problem.

The initial screening ended up with 72 papers. Then we carefully studied through the eligible papers and selected 67 papers that were highly relevant to the topic of our study regarding the limitations and solutions of ADAS based on the topics, methodologies and main findings of the papers. Among the selected papers, 26 of them were related to existing limitations of ADAS (38.8\%), while 41 of the papers were focused on the solutions proposed to improve the performance of ADAS technologies (61.2\%). We then summarized the limitations and solutions of ADAS into three main categories, including human factors, environmental factors, and vehicle factors. Within each main category, subcategories were also defined to further explain the limitations. For the studies related to ADAS limitations, most of them were about “Driver” (40.7\%), followed by “Software” (14.8\%), “Adverse weather” (11.1\%), “Hardware” (11.1\%),  “Lane” (7.4\%), and the rest were “Other driver/vehicle” (3.7\%), “VRU” (3.7\%), “Light condition” (3.7\%), and “Road” (3.7\%). For the studies related to ADAS solutions, the majority were about “Driver” (17.4\%) and “Lane” (17.4\%),  followed by “Hardware” (15.2\%), “Other driver/vehicle” (10.9\%), “Light” (10.9\%), “Software” (10.9\%), and the rest solutions were about “VRU” (8.7\%), “Road” (6.5\%), and “Adverse weather” (2.2\%). Though having a more detailed classification would provide further insight for understanding the limitations and solutions of ADAS, the existing literature did not offer enough information for further analysis.


\subsection{ADAS Limitations}
First, for ADAS limitations related to human factors, we identified three subcategories, i.e., drivers, pedestrians, and other vehicles/drivers. The first subcategory related to human factors was driver misuse or disuse of ADAS, either intentional or unintentional, due to the trust level of the driver in the system and/or the driver's misunderstanding of the system. Example studies \cite{cho2020more,abraham2017autonomous,crump2016differing} showed that drivers’ partial understanding or misunderstanding of the system’s capabilities and limitations, led to system misuse with possible serious consequences. A survey conducted by McDonald et al. \cite{mcdonald2018vehicle} showed that only 21\% of the respondents correctly pointed out that BSW was not designed for extremely high speed, and 62\% believed that FCW would automatically apply brakes. In the meantime, an appropriate level of trust in the system was essential to fully benefit from ADAS. Overtrust caused unintentional misuse as drivers believed that the technology was more capable than it was. For example, drivers engaged in other activities while the ACC was on and did not visually check the blind spot monitor (BSM) when overtrusting the ADAS capability \cite{mcdonald2018vehicle}. The disuse of ADAS was an intentional disengagement of the system when drivers have insufficient trust of the system or when they were annoyed or distracted by system alarms \cite{parasuraman1997humans,jin2019mining}. For example, Reagan and McCartt \cite{reagan2016observed} showed that the LDW function was turned off in 67.2\% of the vehicles observed due to lack of trust and excessive false alarms. 
The second subcategory of human factors is pedestrian. A study led by AAA \cite{american2019automatic} revealed that the functioning of the AEB system was unsuccessful under certain circumstances, such as children darting out, two pedestrians alongside the roadway when their silhouette was unclear from the background, or immediately after a right turn. 
The third category included other drivers and vehicles cutting in the front, causing the ACC system to malfunction with either late braking or excessive braking \cite{chen2021adaptive}. 

The second major limitation of ADAS is that its performance was largely influenced by the environment. Such factors include adverse weather, lighting conditions, and lane and road conditions. It was reported that when the precipitation of rainfall was higher than 20mm/h, the sensors of ADAS would lose effectiveness \cite{goodin2019predicting}. Other adverse weather conditions, such as snow, fog, ice, crosswind, could also block and disable the sensors of ADAS \cite{roh2020analysis}. Extremely low or high temperature was another major cause of sensor failures \cite{capallera2019owner}. Apart from adverse weather, lighting conditions could also offset the benefits of ADAS. While darkness was a common problem, system failures could also be caused by the glare of sunrise or sunset \cite{wang2018efficient}. Many lane related ADAS, such as LKA and LDW might be ineffective under certain road conditions, e.g., when the lane divisions were not clear or covered by objects like snow, leaves, or debris \cite{capallera2019owner}. Moreover, if the road was wet and/or slippery, some ADAS functions, such as ACC and LKA, might be disabled due to the interactions between the tires and the roads \cite{wang2018efficient,wielitzka2017state}.

The third type of ADAS limitations related to the vehicle can be further divided into hardware and software issues. For the hardware part of the system, the limitations of the sensors, such as radars, lidars, and cameras, were the major causes of system malfunctions \cite{wang2018efficient,bmwx3}. The performance of ADAS sensors could also be reduced if vehicle modifications were not compatible with the system. For example, the ADAS sensors might be damaged or obstructed by the trailers attached to the vehicle \cite{capallera2019owner}. For the software issues, cybersecurity was one of the main concerns regarding the ADAS for that ADAS functions could be disturbed by cyberattacks, such as denial-of-service attacks or theft of confidential information \cite{moller2018challenges}. Furthermore, the reliability of algorithms related to sensor fusion, data merging, and other functions was another challenge for ADAS. The information provided by multiple sensors based on sensor fusion algorithms might be inconsistent or completely opposite, which could cause inaccurate and unreliable predictions.  \cite{ziebinski2016survey,kang2020camera}.

\subsection{ADAS Solutions}

Many reviewed studies proposed solutions for the identified ADAS problems, with an overall goal to improve the performance of the system. This section summarizes the solutions based on the same classifications used for limitations. The solutions are categorized into three main categories, human factors, environmental factors, and vehicle factors. We examined whether the ADAS solutions covered all the ADAS limitations identified in the previous studies. For the sub-categories “Light condition”, “Lane”, “Road” and “Hardware”, all the topics described in the limitation-related studies were covered in the solution-related studies, which indicates that the problems were widely studied. However, for the sub-categories “Driver”, “VRU”, “Adverse Weather” and “Software”, only part of the topics were found in the solution-related studies. Among the studies related to human factors, the limitations regarding driver’s trust required were in lack of solutions. And there was an unsolved difficulty caused by silhouette issues of pedestrians walking together. The solutions found to deal with environmental issues were more focused on improving the detection accuracy under light-insufficient conditions, however, the adverse weather related issues such as extreme temperature, snow/ice blockage were also in need of solutions. In the area of software-related limitations of ADAS, the malfunction caused by unreliable and unstable algorithms was in lack of solutions apart from the cyber attack problems. These limitations may need further explorations and investigation to provide a more promising performance of ADAS.

First, the proposed solutions focused on the algorithms and methods for detection and prediction improvement to solve human factor issues. In the subcategory of "Driver", for the limitations related to driver's unintentional misuse caused by inaccurate understanding or overtrust, one solution was to monitor the driver's behaviour, identify, and correct the misbehavior in time. Therefore, researchers proposed algorithms to detect and predict the intentions of drivers to alert them of risky traffic situations and make decisions about whether to take over the driver’s control of the vehicle  \cite{huang2021carpal,deng2020online,khairdoost2020real}. Other studies also examined the methods to improve driver's behavior modeling and detection, such as eye gaze detection, fatigue detection, and other driving behavior detection for better monitoring performance \cite{yuan2021novel,zhao2022research,zhou2020driving,wang2020driver,zheng2020investigation,zhoupredicting2022,zhou2020driver}. In terms of drivers’ intentional disuse of the system, Meuller et al. \cite{mueller2021addressing} proposed a driver monitoring system based on driver behavior tracking to detect deliberate disengagement and misuse of ADAS, the system then used attention reminders and proactive methods to keep drivers engaged in the driver-system interactions and maintain the system's functionalities. 
Pedestrian detection was another major issue for ADAS, and previous studies focused on the framework of intention recognition \cite{piccoli2020fussi}. For example,  studies were proposed to deal with the problem that the pedestrians on the roads were not sensitive enough for the sensors to detect them when they were partially obstructed \cite{castelino2020improving,chen2020integrating}. Other vehicles around the ego vehicle also influenced the ADAS performance. Hence, previous studies were proposed to estimate the location or velocity of surrounding vehicles using sensor fusion \cite{zhu2019fusion,kotur2021camera}, detect vehicle cutting in, lane change intention, and overtaking to avoid accidents \cite{kim2021lane,lin2020vision}.

Second, to address the ADAS failures caused by environmental factors, some researchers proposed machine learning models and algorithms to assist the sensors to make reasonable predictions. For example, Tumas et al. \cite{tumas2021augmentation} proposed a deep learning model to fix sensor distortion under severe weather conditions. Studies also aimed at classifying the road surface condition in order to improve the performance of Electronic Stability Control (ESC), ACC, and AEB systems \cite{lee2021intelligent,vsabanovivc2020identification}. Han et al. \cite{han2020using} developed a deep learning segmentation algorithm to recognize irregular road areas. In order to improve lane detection, many researchers proposed algorithms to deal with unclear or missing lane markings \cite{seo2021study,zhang2021rs,teo2021innovative,khan2020lane}. Teo et al. \cite{teo2021innovative} proposed a Gabor filtering-based algorithm that efficiently detected fading lane boundaries. Studies were also proposed to improve the robustness of lane detection systems under various scenarios, such as rainy or poor lighting conditions \cite{tian2021lsd,chen2019lane,li2020lane}. Another important stream of work was to improve algorithm performance in poor light conditions, such as during nighttime or in areas with insufficient light. Such deep learning algorithms and methods have been utilized to improve the system performance for pedestrian detection, vehicle detection, rear lamp tracking, and so on \cite{nataprawira2021pedestrian,huang2021nighttime,sun2021evolutionary}.

Third, the solutions proposed to deal with the ADAS problems related to vehicle factors can be categorized into “Hardware” and “Software”. From the hardware perspective, the literature were trying to improve the performance of sensors by utilizing sensor fusion and other imaging techniques. Gao et al. \cite{9465646} proposed a radar-imaging technique (MIMO-SAR) to improve the resolution problems from the side view. Another technique related to millimeter-wave frequency modulated continuous wave radars was proposed with an adaptive interference detection and suppression scheme, which could improve the signal to noise ratio under multiple interference environments \cite{9266712}. Sensor fusion is another commonly applied approach to achieve more accurate and reliable performances of sensors. Kotur et al. \cite{9499281} designed deep learning methods to integrate LiDAR points with camera images to improve the accuracy of collision detection and reduce the time for calculation. An coordinate calibration algorithm was proposed by Kim et al \cite{8436959}. to fuse the radar data and vision data, and a better object detection result could be achieved. From the perspective of software, the goal of the solutions was focusing on improve the ADAS performance by software methods and algorithms. Vermiglio et al. \cite{2020} proposed an approach to reduce the computational cost in the system by using reduced order modeling. To deal with possible Sybil attack faced by modern vehicles equipped with ADAS sensors, Lim et al. \cite{9045356} designed a scheme of multi-step verification process to improve the performance of sybil attack detection.






\section{Cause-and-Effect Analysis}

\subsection{Datasets}
We used the dataset from the NHTSA Office of Defects Investigation (ODI) Consumer Complaint database \cite{NHTSA2011}, which includes records of all safety-relevant complaints about vehicles received by NHTSA since January 1, 1995. Complaints included in this project were between October 27, 2015 and May 1, 2021. As for the manufacturer name, the complaints were related to the vehicle manufacturers such as BMW, Mercedes, Porsche, GM, Ford, Honda, Toyota, KIA, Hyundai, Nissan, Subaru, Chrysler, Volkswagen, and Mitsubishi. The total number of complaints in the specified range was 440,999. The database consisted of 49 variables detailing vehicle information, general vehicle component conditions, complaint/incident description, and vehicle purchasing history. 
The primary variable utilized in the study was complaint description, which was an account of the complaint in the consumer’s own words. As some complaints described multiple defective components, the database contained duplicate observations only differed by the defective component description variable. 

\subsection{ADAS Classification}

To examine the cause-and-effect relationship in the the ADAS complaints, we applied Bidirectional Encoder Representations from Transformers (BERT) \cite{devlin2018bert}. To label the dataset used to train and test the classifier models, we firstly identified a list of keywords based on the literature review and by manually checking the complaints to identify the most occurring ADAS related keywords. For instance, a complaint was classified as ADAS related if it contained any of the keywords shown in Table \ref{Table:keywords}. The number of ADAS complaints identified was 1141 and the number of non-ADAS complaints was 439,858. The percentage of complaints containing the identified keywords is shown in Table \ref{Table:keywords}. We used a stratified 10-fold cross validation to train and test the classification models.

\begin{table*}[]
\centering
\caption{Percentage of complaints containing the identified keywords}
\small
\label{Table:keywords}
\begin{tabular}{lc}
\hline
Keywords                                        & \multicolumn{1}{l}{Percentage} \\ \hline
automatic braking, emergency braking system, intelligent braking system, forward emergency braking                    & 23.84 \\
adaptive cruise control, super cruise, smart cruise control, dynamic cruise control                                   & 16.56 \\
lane keep assist                                & 16.56                          \\
automatic steering, steering assist             & 12.80                          \\
autopilot, pilot assist                         & 12.36                          \\
forward collision mitigation system, forward collision avoidance, forward collision alert, rear cross traffic warning & 7.89  \\
blind spot assist, blind spot collision warning & 1.49                           \\
pedestrian detection                            & 0.53                           \\
driver monitoring                               & 0.35                           \\
advanced driver assistance system               & 0.18                           \\
automated driving                               & 0.18                           \\ \hline
\end{tabular}
\end{table*}


\textbf{BERT classifier:} BERT achieved state-of-the-art results in text classification tasks. It was pre-trained on a large number of unlabeled data extracted from BooksCorpus (i.e., 800M words) and English Wikipedia (i.e., 2500M words) \cite{devlin2018bert}. The BERT model consists of  transformer encoders (i.e., read the text) and decoders (i.e., produce a prediction). The general architecture of preparing the input data for BERT transformer is summarized as follows \cite{ayoub2021combat}: 1) the complaint sentences were tokenized and passed through a token embedding layer to transform them into a vector representation of fixed dimension (i.e., 768-dimensional vector), 2) to help the model distinguish between two sentences, extra classification [CLS] and separator [SEP] tokens were added to the start and end of the tokenized sentence, 3) a sentence embedding was added to each token, which helped in classifying a text given a pair of input texts, 4) a position embedding was added to each token to indicate its relative position in a sentence using a sinusoidal function, 5) the final input embedding was a summation of the three embeddings. The summed input was passed to the transformer. In this study, we used the PyTorch-Pretrained-BERT library to build the BERT model. Then, we fine-tuned its linear layer and the sigmoid activation to obtain the predictions with the labeled dataset. During the fine-tuning process, Adam optimizer was used with a learning rate of $3\times 10^{-6}$ and a batch size of 4. We fine-tuned the model on the collected dataset for eight epoch. 

For the assessment of the classification models, the performance was evaluated with precision, recall, and F1-score. Precision represents the proportion of positive samples that were correctly classified to the total number of positive predicted samples, i.e., $Precision = \#true\_positive/(\#true\_positive + \#false\_positive)$. Recall represents the positive correctly classified samples to the total number of positive samples (i.e., ADAS data), i.e., $Recall = \#true\_positive/(\#true\_positive + \#false\_negative)$. F1-score represents the harmonic mean of precision and recall, i.e., F1-score $= 2 \times Precision \times Recall / (Precision + Recall)$ \cite{zhou2011affect}.

\subsection{Causal Model Development}

\begin{figure*}[h]
\centering
\includegraphics [width=1\textwidth]{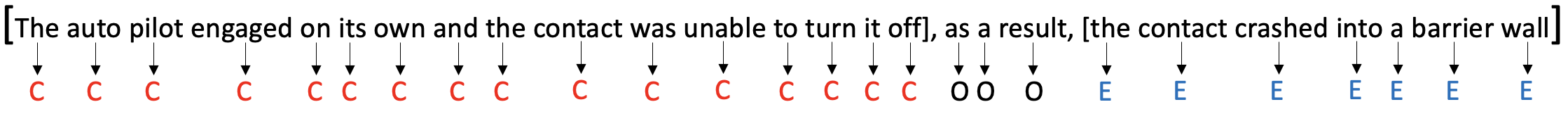}
\caption{Example of cause-and-effect annotation as a sequence labeling problem} 
\label{fig:labeling}
\end{figure*}

In this part of tasks, we annotated the cause-and-effect relationships about ADAS by coding 542 sentences. Out of the selected sentences, 150 complaints with the word “cause”, 150 complaints with the word “due to”, 42 complaints with the words “trigger, lead, result”, and 200 sentences randomly picked. We simplified the cause-and-effect tagging format \cite{zhang2021disengagement}, and the dataset was labeled by three students to increase the reliability of the coding. The inter-raters reliability was 91.5\%. We formulated the problem of cause-and-effect relationship extraction as a sequence labeling problem and we used three labels to annotate a sentence, i.e., O: others, C: Cause, E: Effect (see Figure \ref{fig:labeling}) for an example.

\begin{table}[]
\centering
\caption{ Tags distribution in the dataset}
\label{tab:tag_distribution}
\begin{tabular}{ccc}
\hline
Tags & Meaning      & Number \\ \hline
E    & Effect token & 7161   \\
C    & Cause token  & 5601   \\
O    & Other token  & 73788  \\ \hline
\end{tabular}
\end{table}

In addition to the manually labeled data, we added two publicly available datasets. One was retrieved from the California Department of Motor Vehicles (CA DMV) \cite{californiadmv_2022} and it was labeled by Zhang et al. \cite{zhang2021disengagement} to include cause and effect. The second dataset was a task specific dataset (SemEval-2010 Task 8) \cite{hendrickx2019semeval}. The three datasets (2567 sentences) (see Table 4) were combined then they were used as the ground truth to train the natural language processing model to extract cause-and-effect relations.

Recent developments in natural language processing (NLP) have made substantial progress by using a large amount of unlabeled pre-trained data with a small number of labeled data for specific NLP tasks.  In this work, we used the MPNet model \cite{song2020mpnet}, which is based on permuted language modeling and it uses the auxiliary position information to make the model see a full sentence. Results showed that it performed substantially better than previous models (e.g., BERT \cite{devlin2018bert} and ELECTRA \cite{clark2020electra}) with the same computation budget. Therefore, we fine-tuned the MPNet model with labeled data obtained from the previous step to predict the remaining data. After we constructed the MPNet model, we compared its performance with ELECTRA and BERT models using various performance metrics, including  precision, recall, F1 measure, evaluation loss, and run time.

\section{Results}
\subsection{Classification Results}
A summary of the obtained classification results is shown in Table \ref{tab:adas_classification}. Since the dataset was imbalanced, we varied the number of Non-ADAS complaints. In the first trial, we used the same number of ADAS and Non-ADAS complaints. In the second trial, we doubled the number of Non-ADAS complaints. And in the third trial, we tripled the number of Non-ADAS complaints. A ten-fold-stratified-cross-validation process was used to obtain the results as shown in Table \ref{tab:adas_classification}. The BERT model had slightly better performance when more non-ADAS examples were included, with f1-score = 0.988.

\begin{table}[]
\centering
\caption{ Summary of BERT classifiers performance}
\label{tab:adas_classification}
\begin{tabular}{cccc}
\hline
Sample size                                                           & Precision & Recall & F1-score \\ \hline
ADAS (1141), Non-ADAS (1141) & 0.985     & 0.985  & 0.985    \\
ADAS (1141), Non-ADAS (2282) & 0.986     & 0.992  & \textbf{0.988}    \\
ADAS (1141), Non-ADAS (3423) & 0.985 & 0.992 & 0.988 \\ \hline
\end{tabular}
\end{table}

\subsection{Cause-and-Effect Model Performance and Results}

The performance of the MPNet model was compared to BERT and XLNet since MPNet was built based on these two models. The comparison results of MPNet, ELECTRA, and BERT classification models, including precision, recall, F1 measure, evaluation loss, and run time, are shown in Table \ref{tab:model_performance} using a five-fold cross validation strategy. Precision evaluates the fraction of correctly classified instances among the one classified as positive; it changes from 0 to 1. Recall evaluates the fraction of correctly classified instances among all possible positive predictions and it changes from 0 to 1. F1-score is defined as the harmonic mean between precision and recall. It is the most used metric when learning from imbalanced data, it ranges from 0 to 1. For precision, recall, and F1-score, a score of 1 represents a perfect model.
Evaluation loss or binary cross entropy loss measures how well the classification model performs, it ranges from 0 to 1 with 0 being a perfect model. The MPNet model had a slightly better performance across all metrics than the ELECTRA and BERT models.

\begin{table}[]
\centering
\caption{Summary of model performance}
\label{tab:model_performance}
\begin{tabular}{lccccc}
\hline
Models & \multicolumn{1}{l}{Precision} & \multicolumn{1}{l}{Recall} & \multicolumn{1}{l}{F1-score} & \multicolumn{1}{l}{Loss} & \multicolumn{1}{l}{Run time} \\ \hline
\textbf{MPNet} & 0.838 & 0.847 & \textbf{0.842} & 0.381 & 41 mins \\
Electra        & 0.835 & 0.847 & 0.841          & 0.405 & 46 mins \\
BERT           & 0.852 & 0.847 & 0.839          & 0.571 & 40 mins \\ \hline
\end{tabular}
\end{table}

One example of the predicted tags for the ADAS complaints is shown below: \emph{... tl* the contact owns a 2016 Tesla model s. the contact stated that while driving at 75 mph, the auto pilot engaged on its own as a result, the contact crashed into a barrier wall as he was able to place the vehicle back into manual mode...}\\
Cause: ``the auto pilot engaged on its own and the contact was unable to turn it off ".\\
Effect: ``the contact crashed into a barrier wall".%


\subsection{Taxonomy of Causes and Effects Related to ADAS}

\subsubsection{Categories of ADAS, Causes, and Effects}

Based on the results from the causal model, we manually went through the filtered dataset with 1141 complaints to identify the causes and effects of different ADAS systems. 

First, we identified eight major ADAS categories, including, 1) Emergency braking system (24\%), 2) Adaptive cruise control (23.5\%), 3) Lane keep assist (16.7\%), 4) Automatic steering (12.8\%), 5) Autopilot (12.7\%), 6) Forward collision avoidance (87\%), 7) Blind spot assist (1.5\%), and 8) Others (1.2\%). 
Among them, others included pedestrian detection (0.4\%), driver monitoring (0.2\%), advanced driver assistance system (0.2\%, which did not specify the exact ADAS type), rear cross traffic warning (0.3\%), and super cruise (0.2\%).

In the second step, we identified different categories of causes below: 1) Vehicle factors (71.2\%, including battery, powertrain, wiring issue, data storage issue, over correction, false alarm, windshield crack, automatic engagement, suspension failure, infotainment defect, automatic disengagement, brake failure, over sensitive, main control unit issue, others, recognition error, inattentive driving, electric system issue, steering issue, software issue), 2) Unknown causes (22.6\%), 3) Environment (5.0\%,  road conditions and weather), and 4) human factors (1.2\%, false advertising, other vehicles).

We then examined the detailed categories related to these major categories as shown in Figure \ref{fig:cause_effect}a. The top five causes were unknown (22.6\%), false alarm (17.3\%), fail to respond (13.1\%), recall (6.2\%), and sensor issue (5.5\%). Note when we labeled these categories, we showed the main differences between them below. 
\begin{enumerate}
\item False alarm means that there was nothing on the road, but the vehicle suddenly braked (e.g., \emph{“on multiple occasions while using the adaptive cruise control the vehicle will suddenly and unexpectedly apply the brakes when the path is clear in front of the vehicle”}); 
\item Recognition error means that the vehicle recognizes A as B (e.g., “\emph{while using adaptive cruise control the car will inexplicably break when it sees shadows on the road...}”);
\item Sensor issues mainly include sensor (camera or radar) obstruction or general sensor issues without specification (e.g., “\emph{the adaptive cruise control radar sensor was found crack after being taken to the dealership for error code}”);
\item Fail to respond means that when there were objects with risks to the vehicle, the vehicle’s ADAS did not respond (e.g., “\emph{the forward collision avoidance system was inoperable. the system does not respond to speed limits, does not alarm, does not display pictures, and does not apply any braking force to avoid a crash...}”); 
\item Other vehicles include the driving behavior of others causing the issues (e.g., “\emph{i was driving my 2018 yukon denali on a 3 lane interstate with gm's new adaptive cruise control turned on..  another driver cut in front of me from the right.  adaptive cruise slammed on the brakes almost causing a rear-end accident by overcompensating...} ”);
\item Data storage issues mostly associated with issues of embedded MultiMediaCard (emmc) (e.g., “\emph{the emmc flash memory is failing, leading to the main screen blacking out for several minutes both while driving and when at stop and when the car is booting up. this prevents adjustment of several components including climate (fans, defog, ac), manual lighting, cruise control/autopilot, etc.”)}”);
\item Inattentive driving includes mostly handoff driving from Tesla vehicles (e.g., “\emph{he is not taking the advice to hold the wheel seriously and his placing lives at risk by his reckless behavior...}”);
\item Others include causes that were less than 5 cases, including engine, fuel pump, leaking-control box, leaking-cylinder, leaking-oil, mirror issue, remote start, security design, speed adjuster, switch, transmission problem, unintended acceleration, unintended deceleration, and unintended lane change. 
\end{enumerate}
Thirdly, we identified different categories of effects of these causes as shown in Figure \ref{fig:cause_effect}b. The top five effects were 1) ADAS failure (18.8\%), hard braking (13.0\%), warning alert (10.5\%), collision (9.2\%), and near collision (5.9\%). We also explained the main differences among categories when we labeled them as follows:

\begin{figure*} [h]
	\centering
	\subfloat[\label{fig:causes}]{\includegraphics[width=.8\textwidth]{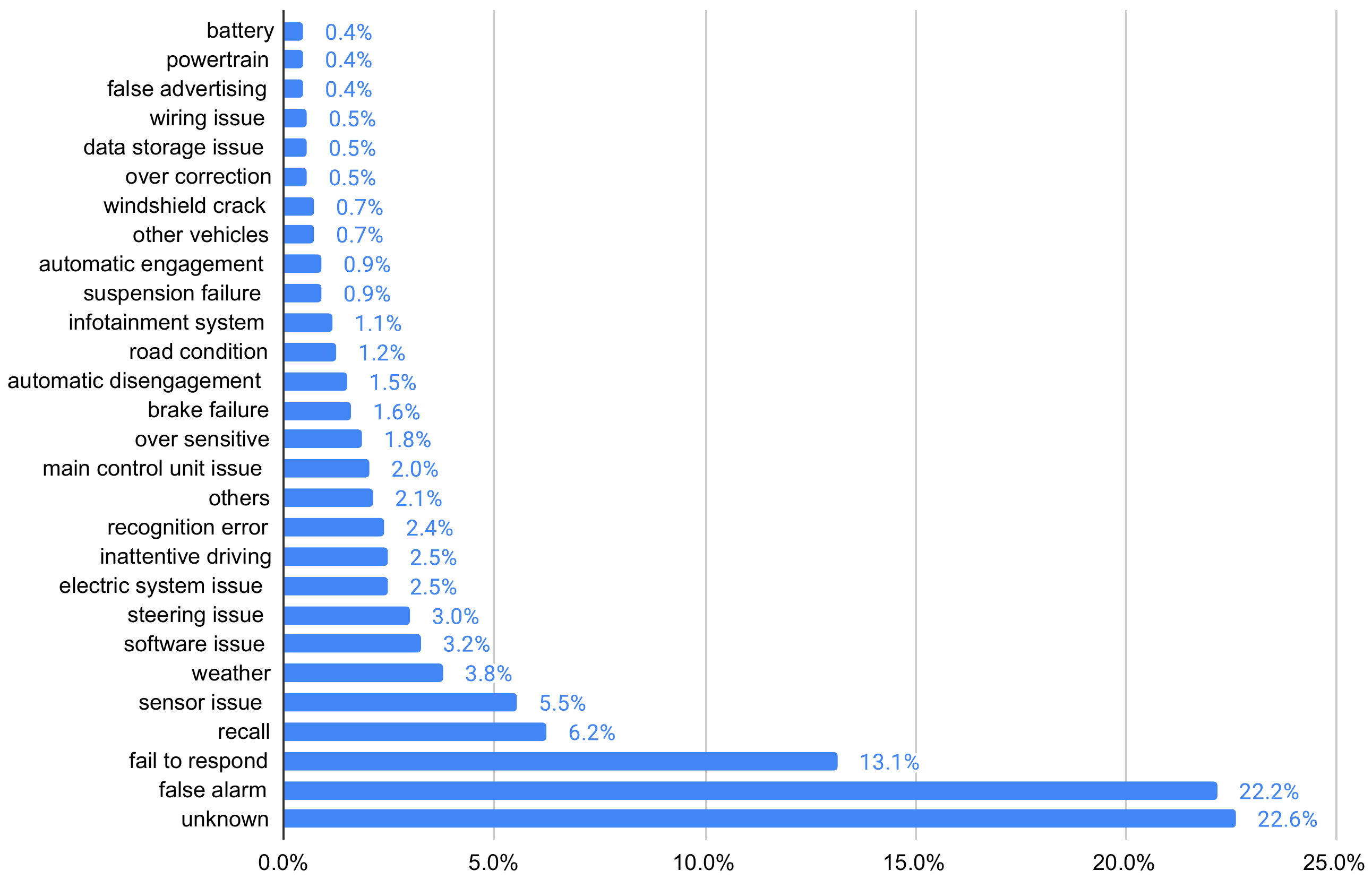}}
	\hspace{0pt}
	\subfloat[\label{fig:effects}]{\includegraphics[width=.8\textwidth]{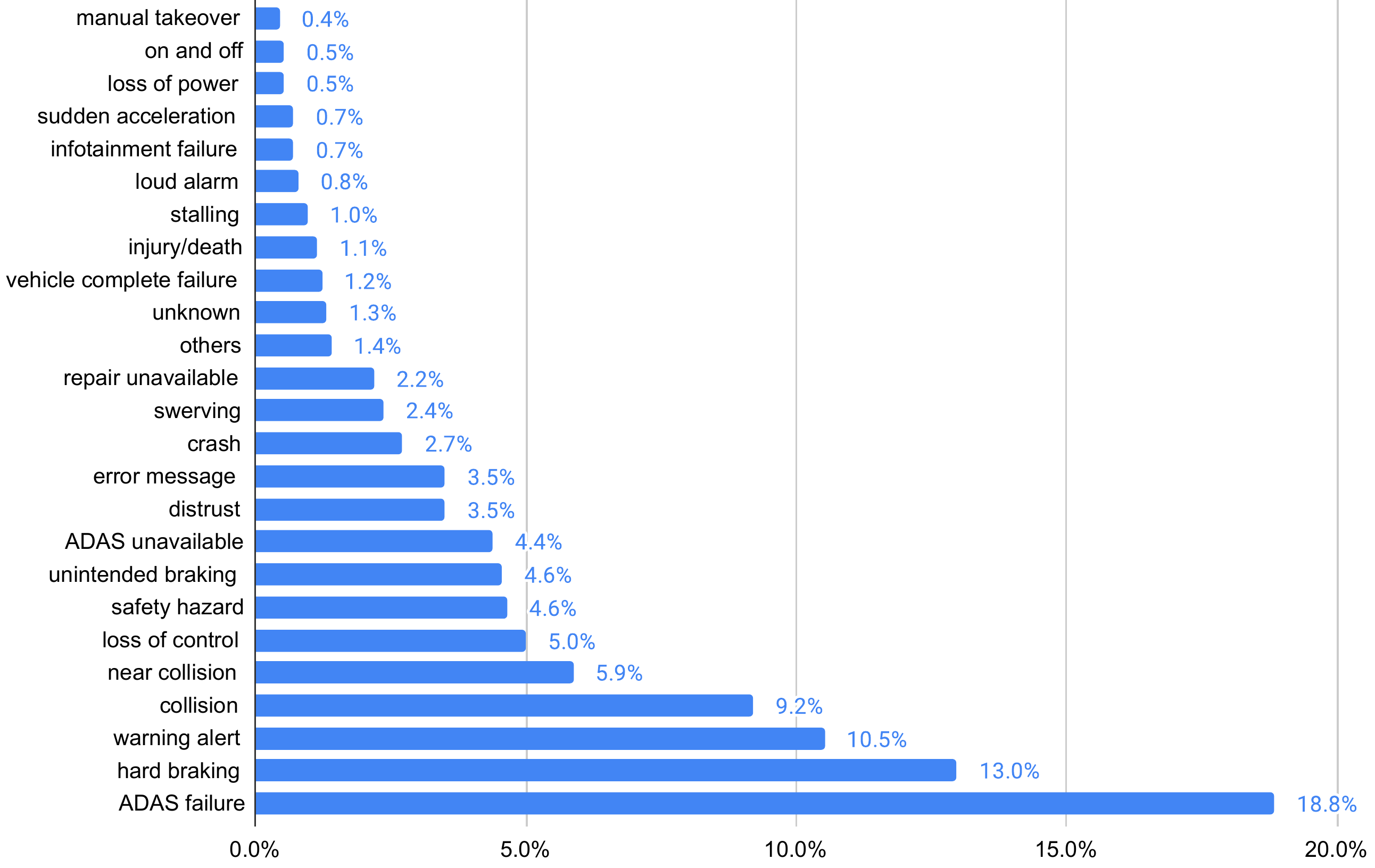}}
    \caption{Percentages of (a) causes and (b) effects of ADAS failures.}\label{fig:cause_effect}
\end{figure*}

\begin{enumerate}
\item Collision was less severe than crash, such as “\emph{i had to merge into traffic from the hov lane disengaging the cruise control. i then used the  auto transmission to shift to a lower gear to gain more torque in changing lanes when the car went into a stall and lost all power. consequently we were rear-ended by another car.}”
\item Crash was more severe than collision, such as “\emph{he credits the tesla with saving his life but does not concede that the crash might have been caused by a vehicle fault... both front and rear wheels have been torn off the car which may have been the result of the crash ...}”
\item Crash with injury or death was the most severe, such as “\emph{...lane assist jerked the wheel as i was approaching the car in front of me and caused me to go into the other lane and and then it resumed speed because it did not realize a vehicle was coming the other way. at this time i had no control of the car. by the time i could get back over and got control of the vehicle it was to late and i hit someone and they died.}”
\item Loss of control is more general from the description than hard braking or swerving which were explicitly mentioned, such as “\emph{...  this even happens when " the lane keep assist" is switched off.  it is a constant struggle to keep it from drifting into the right lane and other cars...  the pull is so strong that it makes the lane keep assist system (lkas) unusable.  when using  lkas on cruise control the vehicle keeps  drifting into the shoulder of the road and even crosses the line.}” 
One example of hard braking was, “\emph{i was driving home... a message on the display stated that a crash was imminent. then the brakes engaged suddenly, and the car almost came to a complete stop in the middle of the street. the brakes engaged so hard and so quickly, that it threw me forward in the seat.}” 
One example of swerving was “\emph{when this happened on highways while using autopilot (experienced at least three times), the car swerved left and right erratically (after changing lanes) in an effort to re-center on the lane without success, and i had to take over manually in over to avoid from flipping over or an accident.}”
\item Stalling (not drivable but with infotainment on) was different from vehicle complete failure which lost all power of the vehicle. One example of stalling was “\emph{vehicle is stalling at lights. vehicle has rough idle. vehicle automatic braking activate without warning.}” 
One example of vehicle complete failure was “\emph{2017 honda crv parked in garage, when attempting to start engine dash lights up warnings on braking system, cruise control, lane assist, battery and many other areas.  car would not start and goes dead with 3rd attempt to start.}”
\item Safety hazard is used when it was explicitly mentioned while unknown did not mention any results of the complaint. One example of safety hazard was “\emph{van has just over 5k miles... primary warning was 'low brake power' and braking became extremely touchy as it seemed to utilize the emergency brake as a backup. this extreme braking is hazardous and could result in an accident caused by braking too quickly - with a possible rear-end collision by traffic to the rear.}”
\end{enumerate}

\subsubsection{Causes to ADAS issues}

By examining the cause and effect pairs associated with each ADAS issue, we were able to identify the most important causes to specific ADAS issues and their associated effects.Table 6 shows the top five causes and top five effects to each ADAS category. For example, the top five causes of forward collision avoidance were recall (41.4\%), false alarm (18.4\%), unknown (14.9\%), fail to respond (11.5\%) and sensor issue (5.7\%), which accounted for 91.9\% of the cases. For example, the top five effects of forward collision avoidance were repair unavailable (27.6\%), hard braking (18.4\%), ADAS failure (11.5\%), warning alert (11.5\%) and unintended braking (8.0\%), which accounted for 77.0\% of the cases. We were also able to identify the top 10 cause-effect pairs by aggregating different ADAS categories in Table \ref{tab:top10causes}.



\begin{table*}[h]
\centering
\caption{Top 10 causes to effects by aggregating different ADAS categories }
\label{tab:top10causes}
\begin{tabular}{cccc}
\hline
\textbf{Cause}      & \textbf{Effect}    & \textbf{\#Complaints} & \textbf{\%Complaints} \\ \hline
unknown             & ADAS failure       & 87                  & 7.6\%                 \\
false alarm         & hard braking       & 85                  & 7.4\%                 \\
fail to respond     & collision          & 46                  & 4\%                   \\
unknown             & warning alert      & 40                  & 3.5\%                 \\
recall              & repair unavailable & 24                  & 2.1\%                 \\
false alarm         & unintended braking & 22                  & 1.9\%                 \\
false alarm         & warning alert      & 21                  & 1.8\%                 \\
inattentive driving & collision          & 19                  & 1.7\%                 \\
unknown             & error message      & 18                  & 1.6\%                 \\
sensor issue        & ADAS failure       & 17                  & 1.5\%                 \\ \hline
\end{tabular}
\end{table*}

\begin{table*}[]
\centering
\caption{Top 5 causes and effects of the identified ADAS categories with their percentages}
\label{tablef}
\scalebox{0.71}{
\begin{tabular}{l 
c 
c }
\hline
\multicolumn{1}{c}{{\color[HTML]{222222} \textbf{ADAS category}}} & {\color[HTML]{222222} \textbf{Top five causes}} & {\color[HTML]{222222} \textbf{Top five effects}} \\ \hline
Forward collision Avoidance & {\color[HTML]{222222} \begin{tabular}[c]{@{}c@{}}Recall (41.4), false alarm (18.4), unknown (14.9), \\ fail to respond (11.5), sensor issue (5.7)\end{tabular}} & {\color[HTML]{222222} \begin{tabular}[c]{@{}c@{}}Repair unavailable (27.6), hard braking (18.4), ADAS failure (11.5),\\ warning alert (11.5), unintended braking (8)\end{tabular}} \\
Emergency braking system & {\color[HTML]{222222} \begin{tabular}[c]{@{}c@{}}False alarm (45.3), fail to respond (13.5), sensor issue (12), \\ unknown (10.9), over sensitive (7.3)\end{tabular}} & {\color[HTML]{222222} \begin{tabular}[c]{@{}c@{}}Hard braking (28.9), warning alert (15.7), collision (11.7), \\ ADAS failure (8.4), unintended braking (7.3)\end{tabular}} \\
Automatic steering & {\color[HTML]{222222} \begin{tabular}[c]{@{}c@{}}Unknown (45.9), recall (15.8), steering issue (11), \\ electric system issue (6.8), weather (5.5)\end{tabular}} & {\color[HTML]{222222} \begin{tabular}[c]{@{}c@{}}ADAS failure (40.4), error message (20.5), warning alert (11.6), \\ loss of control (6.8), ADAS unavailable (4.8)\end{tabular}} \\
Autopilot & {\color[HTML]{222222} \begin{tabular}[c]{@{}c@{}}Fail to respond (21.4), inattentive driving (19.3), unknown (13.1), \\ main control unit issue (11.7), software issue (8.3)\end{tabular}} & {\color[HTML]{222222} \begin{tabular}[c]{@{}c@{}}Collision (31), ADAS unavailable (13.1), ADAS failure (9.7), \\ loss of control (6.9), hard braking (6.2)\end{tabular}} \\
Blind spot assist & {\color[HTML]{222222} \begin{tabular}[c]{@{}c@{}}Unknown (23.5), others (17.6), battery (11.8), \\ automatic disengage (5.9), electric system issue (5.9)\end{tabular}} & {\color[HTML]{222222} \begin{tabular}[c]{@{}c@{}}ADAS failure (41.2), warning alert (17.6), safety hazard (11.8),\\  collision (5.9), error message (5.9)\end{tabular}} \\
Lane keep assist & {\color[HTML]{222222} \begin{tabular}[c]{@{}c@{}}Unknown (24.7), fail to respond (11.6), recognition error (8.9), \\ steering issue (7.9), false alarm (5.8)\end{tabular}} & {\color[HTML]{222222} \begin{tabular}[c]{@{}c@{}}ADAS failure (21.6), loss of control (14.7), swerving (12.1), \\ warning alert (11.6), distrust (6.8)\end{tabular}} \\ \hline
\end{tabular}%
}
\end{table*}

\section{Discussions} 
This study investigated the issues and solutions of ADAS failures in the literature and identified the causes and effects from a consumer complaint dataset. The key findings of both efforts are highlighted below.


\textbf{Major ADAS issues}. We found three main categories of factors that were associated with ADAS issues in both efforts of this study, including human factors, environmental, and vehicle factors. In the literature review, we found that ADAS issues were mostly related to human factors (48.1\%), followed by vehicle factors (26.0\%) and environmental factors (25.9\%). Thus, many solutions were proposed with regard to human factors solutions. However, in the consumer complaint dataset, we found that  71.2\% of the complaints were associated with vehicle, followed by unknown factors (22.6\%), environmental factors (5.0\%), and human factors (1.2\%). The differences in terms of the distributions of causes showed different possible biases in self-reported complaint data that humans tend to complain more about external factors rather than internal factors (i.e., human factors). Even though, human factors issues were reported in the consumer complaint dataset, they were related to false advertising and other vehicles. Nevertheless, we were able to identify a variety of causes of ADAS related to vehicles in the consumer complaint dataset (see Figure \ref{fig:cause_effect}a) with top five causes: false alarm (17.3\%), fail to respond (13.1\%), recall (6.2\%), and sensor issue (5.5\%), and software issues (3.2\%). We found the environmental factors played key roles in leading to ADAS issues, including road and lane conditions, weather and lighting, which seemed to be consistent between the two sources.  At the same time, we were able to identify the leading causes of individual ADAS systems through the consumer complaint dataset (see Table 6), which were not available in the literature. Such findings helped provide more information to understand the possible solutions to improve ADAS more specifically in reality. In this sense, these two lines of research tend to complement each other with a better understanding of the the causes of ADAS.

\textbf{ADAS solutions and effects}. In the literature, researchers were able to provide a variety number of solutions based on the issues identified. A larger number of studies focused on developing advanced algorithms to improve the performance of ADAS that were able to improve driver trust and less vulnerable to different weather and road conditions (e.g., \cite{huang2021carpal,deng2020online,khairdoost2020real}). Another important finding was that researchers were also interested in modeling and monitoring driver behavior to help improve the joint performance of the human-vehicle system (e.g., \cite{yuan2021novel,zhao2022research,zhou2020driving}.
However, in the consumer complaint dataset, we emphasized the possible consequences of the ADAS issues despite the fact that there were a small number of solutions to ADAS issues provided by dealers, mechanics, or automotive companies, but these might be temporary solutions since many ADAS problems were still persistent (e.g., \emph{tl* the contact owns a 2019 honda cr-v. while the contact was driving approximately 45 mph, another vehicle pausing on the passenger side of the contact's vehicle caused the automatic braking system to activate. as a result, the contact's vehicle independently and abruptly stopped... the vehicle was taken to first team Honda where it was diagnosed that the forward collision avoidance sensor lens was contaminated with road debris. the lens was cleaned and the vehicle was returned to the contact. the contact stated that the failure recurred within thirty minutes...}). 
The effects of ADAS issues identified were either associated with specific ADAS or specific causes. For example, Table 6 shows the top effects related to specific ADAS while Table 5 shows the most frequently cause and effect pairs in the dataset. These insights provided guideline in prioritization in terms of fixing different issues associated with ADAS. For example, distrust was one of the top 5 effects associated with lane keep assist, which indicated an urgent need to improve the performance of such ADAS in different scenarios.

\textbf{Natural language processing models}. In this study, we made use of the state-of-the-art pre-trained language models to perform two important tasks, i.e., ADAS classification and cause-and-effect relation extraction. We used BERT to conduct ADAS classification. The model performed well for this task, which helped us identify the ADAS related consumer complaints. Then we used three different types of language models for cause-and-effect relation extraction and it was found that MPNet performed slightly better than ELECTRA and BERT. This task was much more difficult than ADAS classification. We used labeled dataset for fine-tuning the language models. However, the labeled datasets were from different domains. More research is needed to further improve the performance of the cause-and-effect extraction.

\textbf{Limitations and future work}. The study has some limitations that can be addressed in future studies. 
First, the findings from the literature review and the consumer complaint dataset might be biased toward specific causes of ADAS issues. The literature review was fairly objective by focusing the current major ADAS limitations and their potential solutions from research point of view, but might ignore other issues that would happen in our daily life.  The consumer complaint dataset was more subjective and tended to be biased toward vehicle factors associated with ADAS causes.
Thanks to the nature of these two sources, the results were complementary to each other. Future research can be conducted to examine other resources to identify the causes of ADAS issues and solutions more comprehensively, such as patents. 
Second, due to the cost associated with labeling cause-and-effect relations, we only had a small labeled dataset to fine-tune the pre-trained language models to extract causes and effects of ADAS in the consumer complaint dataset. Although our MPNet was able to identify the causes and effects with f1-score = 0.842, more work is needed to further improve the performance of cause and effect extraction.

\section{Conclusion}
In this study, we conducted a comprehensive literature review on the causes and solutions of ADAS failure and we found that both the causes and solutions of ADAS can be categorized into human factors, environmental factors, and vehicle factors. Then we conducted a cause-effect analysis with an NHTSA ODI Consumer Complaint database using natural language processing techniques based on deep learning models. Our model was able to identify causes and effects with f1-score = 84.1\%, based on which we identified the causes and effects associated with different ADAS categories. At the same time, we identified 1) top 5 causes of each ADAS, 2) top 5 effects of each ADAS, and 3) top 10 most frequent cause and effect pairs. Then the findings from these two efforts were compared and discussed to provide possible recommendations to improve the safety of ADAS.

\begin{acks}
This work was supported by TOYOTA MOTOR ENGINEERING and MANUFACTURING, NORTH AMERICA, INC.
\end{acks}


\bibliographystyle{ACM-Reference-Format}
\bibliography{sample-base}


\begin{thebibliography}{78}


\ifx \showCODEN    \undefined \def \showCODEN     #1{\unskip}     \fi
\ifx \showDOI      \undefined \def \showDOI       #1{#1}\fi
\ifx \showISBNx    \undefined \def \showISBNx     #1{\unskip}     \fi
\ifx \showISBNxiii \undefined \def \showISBNxiii  #1{\unskip}     \fi
\ifx \showISSN     \undefined \def \showISSN      #1{\unskip}     \fi
\ifx \showLCCN     \undefined \def \showLCCN      #1{\unskip}     \fi
\ifx \shownote     \undefined \def \shownote      #1{#1}          \fi
\ifx \showarticletitle \undefined \def \showarticletitle #1{#1}   \fi
\ifx \showURL      \undefined \def \showURL       {\relax}        \fi
\providecommand\bibfield[2]{#2}
\providecommand\bibinfo[2]{#2}
\providecommand\natexlab[1]{#1}
\providecommand\showeprint[2][]{arXiv:#2}

\bibitem[\protect\citeauthoryear{??}{cal}{2022}]%
        {californiadmv_2022}
 \bibinfo{year}{2022}\natexlab{}.
\newblock \bibinfo{title}{Autonomous vehicles}.
\newblock
\newblock
\urldef\tempurl%
\url{https://www.dmv.ca.gov/portal/vehicle-industry-services/autonomous-vehicles/}
\showURL{%
\tempurl}


\bibitem[\protect\citeauthoryear{Abraham, Lee, Brady, Fitzgerald, Mehler,
  Reimer, and Coughlin}{Abraham et~al\mbox{.}}{2017}]%
        {abraham2017autonomous}
\bibfield{author}{\bibinfo{person}{Hillary Abraham}, \bibinfo{person}{Chaiwoo
  Lee}, \bibinfo{person}{Samantha Brady}, \bibinfo{person}{Craig Fitzgerald},
  \bibinfo{person}{Bruce Mehler}, \bibinfo{person}{Bryan Reimer}, {and}
  \bibinfo{person}{Joseph~F Coughlin}.} \bibinfo{year}{2017}\natexlab{}.
\newblock \showarticletitle{Autonomous vehicles and alternatives to driving:
  trust, preferences, and effects of age}. In
  \bibinfo{booktitle}{\emph{Proceedings of the transportation research board
  96th annual meeting (TRB'17)}}.
\newblock


\bibitem[\protect\citeauthoryear{Association et~al\mbox{.}}{Association
  et~al\mbox{.}}{2019}]%
        {american2019automatic}
\bibfield{author}{\bibinfo{person}{American~Automobile Association}
  {et~al\mbox{.}}} \bibinfo{year}{2019}\natexlab{}.
\newblock \showarticletitle{Automatic emergency braking with pedestrian
  detection}.
\newblock \bibinfo{journal}{\emph{Retrieved January}}  \bibinfo{volume}{20}
  (\bibinfo{year}{2019}), \bibinfo{pages}{2020}.
\newblock
\urldef\tempurl%
\url{https://www.aaa.com/AAA/common/aar/files/Research-Report-Pedestrian-Detection.pdf}
\showURL{%
\tempurl}


\bibitem[\protect\citeauthoryear{Ayoub, Avetisyan, Makki, and Zhou}{Ayoub
  et~al\mbox{.}}{2021a}]%
        {ayoub2021investigation}
\bibfield{author}{\bibinfo{person}{Jackie Ayoub}, \bibinfo{person}{Lilit
  Avetisyan}, \bibinfo{person}{Mustapha Makki}, {and} \bibinfo{person}{Feng
  Zhou}.} \bibinfo{year}{2021}\natexlab{a}.
\newblock \showarticletitle{An Investigation of Drivers’ Dynamic Situational
  Trust in Conditionally Automated Driving}.
\newblock \bibinfo{journal}{\emph{IEEE Transactions on Human-Machine Systems}}
  (\bibinfo{year}{2021}).
\newblock


\bibitem[\protect\citeauthoryear{Ayoub, Yang, and Zhou}{Ayoub
  et~al\mbox{.}}{2021b}]%
        {ayoub2021combat}
\bibfield{author}{\bibinfo{person}{Jackie Ayoub}, \bibinfo{person}{X~Jessie
  Yang}, {and} \bibinfo{person}{Feng Zhou}.} \bibinfo{year}{2021}\natexlab{b}.
\newblock \showarticletitle{Combat COVID-19 infodemic using explainable natural
  language processing models}.
\newblock \bibinfo{journal}{\emph{Information Processing \& Management}}
  \bibinfo{volume}{58}, \bibinfo{number}{4} (\bibinfo{year}{2021}),
  \bibinfo{pages}{102569}.
\newblock
\urldef\tempurl%
\url{https://doi.org/10.1016/j.ipm.2021.102569}
\showDOI{\tempurl}


\bibitem[\protect\citeauthoryear{Ayoub, Yang, and Zhou}{Ayoub
  et~al\mbox{.}}{2021c}]%
        {ayoub2021modeling}
\bibfield{author}{\bibinfo{person}{Jackie Ayoub}, \bibinfo{person}{X~Jessie
  Yang}, {and} \bibinfo{person}{Feng Zhou}.} \bibinfo{year}{2021}\natexlab{c}.
\newblock \showarticletitle{Modeling dispositional and initial learned trust in
  automated vehicles with predictability and explainability}.
\newblock \bibinfo{journal}{\emph{Transportation research part F: traffic
  psychology and behaviour}}  \bibinfo{volume}{77} (\bibinfo{year}{2021}),
  \bibinfo{pages}{102--116}.
\newblock


\bibitem[\protect\citeauthoryear{BMW}{BMW}{[n.d.]}]%
        {bmwx3}
\bibfield{author}{\bibinfo{person}{BMW}.} \bibinfo{year}{[n.d.]}\natexlab{}.
\newblock \bibinfo{booktitle}{\emph{BMW X3 — Owner's Manual}}.
\newblock BMW.
\newblock


\bibitem[\protect\citeauthoryear{Capallera, Meteier, de~Salis, Angelini,
  Carrino, Khaled, and Mugellini}{Capallera et~al\mbox{.}}{2019}]%
        {capallera2019owner}
\bibfield{author}{\bibinfo{person}{Marine Capallera}, \bibinfo{person}{Quentin
  Meteier}, \bibinfo{person}{Emmanuel de Salis}, \bibinfo{person}{Leonardo
  Angelini}, \bibinfo{person}{Stefano Carrino}, \bibinfo{person}{Omar~Abou
  Khaled}, {and} \bibinfo{person}{Elena Mugellini}.}
  \bibinfo{year}{2019}\natexlab{}.
\newblock \showarticletitle{Owner manuals review and taxonomy of ADAS
  limitations in partially automated vehicles}. In
  \bibinfo{booktitle}{\emph{Proceedings of the 11th International Conference on
  Automotive User Interfaces and Interactive Vehicular Applications}}.
  \bibinfo{pages}{156--164}.
\newblock
\urldef\tempurl%
\url{https://doi.org/10.1145/3342197.3344530}
\showDOI{\tempurl}


\bibitem[\protect\citeauthoryear{Castelino, Pinheiro, Praciano, Santos,
  Weichenberger, and J{\'u}nior}{Castelino et~al\mbox{.}}{2020}]%
        {castelino2020improving}
\bibfield{author}{\bibinfo{person}{Redge~Melroy Castelino},
  \bibinfo{person}{Gabriel Passos~Moreira Pinheiro}, \bibinfo{person}{Bruno
  Justino~Garcia Praciano}, \bibinfo{person}{Giovanni~Almeida Santos},
  \bibinfo{person}{Lothar Weichenberger}, {and} \bibinfo{person}{Rafael
  Tim{\'o}teo De~Sousa J{\'u}nior}.} \bibinfo{year}{2020}\natexlab{}.
\newblock \showarticletitle{Improving the Accuracy of Pedestrian Detection in
  Partially Occluded or Obstructed Scenarios}. In
  \bibinfo{booktitle}{\emph{2020 10th International Conference on Advanced
  Computer Information Technologies (ACIT)}}. IEEE, \bibinfo{pages}{834--838}.
\newblock
\urldef\tempurl%
\url{https://doi.org/10.1109/ACIT49673.2020.9208877}
\showDOI{\tempurl}


\bibitem[\protect\citeauthoryear{Chen, Guo, Guo, Chen, Zhang, and Wang}{Chen
  et~al\mbox{.}}{2021}]%
        {chen2021adaptive}
\bibfield{author}{\bibinfo{person}{Chongpu Chen}, \bibinfo{person}{Jianhua
  Guo}, \bibinfo{person}{Chong Guo}, \bibinfo{person}{Chaoyi Chen},
  \bibinfo{person}{Yao Zhang}, {and} \bibinfo{person}{Jiawei Wang}.}
  \bibinfo{year}{2021}\natexlab{}.
\newblock \showarticletitle{Adaptive Cruise Control for Cut-In Scenarios Based
  on Model Predictive Control Algorithm}.
\newblock \bibinfo{journal}{\emph{Applied Sciences}} \bibinfo{volume}{11},
  \bibinfo{number}{11} (\bibinfo{year}{2021}), \bibinfo{pages}{5293}.
\newblock
\urldef\tempurl%
\url{https://doi.org/10.3390/app11115293}
\showDOI{\tempurl}


\bibitem[\protect\citeauthoryear{Chen, Lin, Dai, and Chen}{Chen
  et~al\mbox{.}}{2015}]%
        {chen2015typical}
\bibfield{author}{\bibinfo{person}{Qiang Chen}, \bibinfo{person}{Miao Lin},
  \bibinfo{person}{Bing Dai}, {and} \bibinfo{person}{Jiguang Chen}.}
  \bibinfo{year}{2015}\natexlab{}.
\newblock \showarticletitle{Typical pedestrian accident scenarios in China and
  crash severity mitigation by autonomous emergency braking systems}. In
  \bibinfo{booktitle}{\emph{SAE 2015 World Congress Proceedings. SAE
  International}}.
\newblock
\urldef\tempurl%
\url{https://doi.org/10.4271/2015-01-1464.}
\showDOI{\tempurl}


\bibitem[\protect\citeauthoryear{Chen, Li, Guo, and Zhou}{Chen
  et~al\mbox{.}}{2019}]%
        {chen2019lane}
\bibfield{author}{\bibinfo{person}{Shizeng Chen}, \bibinfo{person}{Bijun Li},
  \bibinfo{person}{Yuan Guo}, {and} \bibinfo{person}{Jian Zhou}.}
  \bibinfo{year}{2019}\natexlab{}.
\newblock \showarticletitle{Lane Detection Based on Histogram of Oriented
  Vanishing Points}. In \bibinfo{booktitle}{\emph{Asian Conference on Pattern
  Recognition}}. Springer, \bibinfo{pages}{3--11}.
\newblock
\urldef\tempurl%
\url{https://doi.org/10.1007/978-981-15-3651-9_1}
\showDOI{\tempurl}


\bibitem[\protect\citeauthoryear{Chen, Kuan, and Chiang}{Chen
  et~al\mbox{.}}{2020}]%
        {chen2020integrating}
\bibfield{author}{\bibinfo{person}{Wen-Hui Chen}, \bibinfo{person}{Chi-Wei
  Kuan}, {and} \bibinfo{person}{Chuan-Cho Chiang}.}
  \bibinfo{year}{2020}\natexlab{}.
\newblock \showarticletitle{Integrating Multiscale Deformable Part Models and
  Convolutional Networks for Pedestrian Detection.}. In
  \bibinfo{booktitle}{\emph{VEHITS}}. \bibinfo{pages}{515--521}.
\newblock


\bibitem[\protect\citeauthoryear{Cho and Heo}{Cho and Heo}{2020}]%
        {cho2020more}
\bibfield{author}{\bibinfo{person}{Jiyong Cho} {and} \bibinfo{person}{Jeongyun
  Heo}.} \bibinfo{year}{2020}\natexlab{}.
\newblock \showarticletitle{The More You Know, The More You Can Trust:
  Drivers’ Understanding of the Advanced Driver Assistance System}. In
  \bibinfo{booktitle}{\emph{International Conference on Human-Computer
  Interaction}}. Springer, \bibinfo{pages}{230--248}.
\newblock
\urldef\tempurl%
\url{https://doi.org/10.1007/978-3-030-50523-3_16}
\showDOI{\tempurl}


\bibitem[\protect\citeauthoryear{Cicchino}{Cicchino}{2017}]%
        {cicchino2017effectiveness}
\bibfield{author}{\bibinfo{person}{Jessica~B Cicchino}.}
  \bibinfo{year}{2017}\natexlab{}.
\newblock \showarticletitle{Effectiveness of forward collision warning and
  autonomous emergency braking systems in reducing front-to-rear crash rates}.
\newblock \bibinfo{journal}{\emph{Accident Analysis \& Prevention}}
  \bibinfo{volume}{99} (\bibinfo{year}{2017}), \bibinfo{pages}{142--152}.
\newblock
\urldef\tempurl%
\url{https://doi.org/10.1016/j.aap.2016.11.009}
\showDOI{\tempurl}


\bibitem[\protect\citeauthoryear{Clark, Luong, Le, and Manning}{Clark
  et~al\mbox{.}}{2020}]%
        {clark2020electra}
\bibfield{author}{\bibinfo{person}{Kevin Clark}, \bibinfo{person}{Minh-Thang
  Luong}, \bibinfo{person}{Quoc~V Le}, {and} \bibinfo{person}{Christopher~D
  Manning}.} \bibinfo{year}{2020}\natexlab{}.
\newblock \showarticletitle{Electra: Pre-training text encoders as
  discriminators rather than generators}.
\newblock \bibinfo{journal}{\emph{arXiv preprint arXiv:2003.10555}}
  (\bibinfo{year}{2020}).
\newblock
\urldef\tempurl%
\url{https://doi.org/10.48550/arXiv.2003.10555}
\showDOI{\tempurl}


\bibitem[\protect\citeauthoryear{Crump, Cades, Lester, Reed, Barakat, Milan,
  and Young}{Crump et~al\mbox{.}}{2016}]%
        {crump2016differing}
\bibfield{author}{\bibinfo{person}{Caroline Crump}, \bibinfo{person}{David
  Cades}, \bibinfo{person}{Benjamin Lester}, \bibinfo{person}{Scott Reed},
  \bibinfo{person}{Brandon Barakat}, \bibinfo{person}{Laurene Milan}, {and}
  \bibinfo{person}{Douglas Young}.} \bibinfo{year}{2016}\natexlab{}.
\newblock \showarticletitle{Differing perceptions of advanced driver assistance
  systems (ADAS)}. In \bibinfo{booktitle}{\emph{Proceedings of the Human
  Factors and Ergonomics Society Annual Meeting}}, Vol.~\bibinfo{volume}{60}.
  SAGE Publications Sage CA: Los Angeles, CA, \bibinfo{pages}{861--865}.
\newblock
\urldef\tempurl%
\url{https://doi.org/10.1177/1541931213601197}
\showDOI{\tempurl}


\bibitem[\protect\citeauthoryear{Deng, Saleh, Tanshi, and S{\"o}ffker}{Deng
  et~al\mbox{.}}{2020}]%
        {deng2020online}
\bibfield{author}{\bibinfo{person}{Qi Deng}, \bibinfo{person}{Maryam Saleh},
  \bibinfo{person}{Foghor Tanshi}, {and} \bibinfo{person}{Dirk S{\"o}ffker}.}
  \bibinfo{year}{2020}\natexlab{}.
\newblock \showarticletitle{Online intention recognition applied to real
  simulated driving maneuvers}. In \bibinfo{booktitle}{\emph{2020 IEEE
  Conference on Cognitive and Computational Aspects of Situation Management
  (CogSIMA)}}. IEEE, \bibinfo{pages}{1--6}.
\newblock
\urldef\tempurl%
\url{https://doi.org/10.1109/CogSIMA49017.2020.9216115}
\showDOI{\tempurl}


\bibitem[\protect\citeauthoryear{Devlin, Chang, Lee, and Toutanova}{Devlin
  et~al\mbox{.}}{2018}]%
        {devlin2018bert}
\bibfield{author}{\bibinfo{person}{Jacob Devlin}, \bibinfo{person}{Ming-Wei
  Chang}, \bibinfo{person}{Kenton Lee}, {and} \bibinfo{person}{Kristina
  Toutanova}.} \bibinfo{year}{2018}\natexlab{}.
\newblock \showarticletitle{Bert: Pre-training of deep bidirectional
  transformers for language understanding}.
\newblock \bibinfo{journal}{\emph{arXiv preprint arXiv:1810.04805}}
  (\bibinfo{year}{2018}).
\newblock
\urldef\tempurl%
\url{https://doi.org/10.48550/arXiv.1810.04805}
\showDOI{\tempurl}


\bibitem[\protect\citeauthoryear{Gao, Roy, and Xing}{Gao et~al\mbox{.}}{2021}]%
        {9465646}
\bibfield{author}{\bibinfo{person}{Xiangyu Gao}, \bibinfo{person}{Sumit Roy},
  {and} \bibinfo{person}{Guanbin Xing}.} \bibinfo{year}{2021}\natexlab{}.
\newblock \showarticletitle{MIMO-SAR: A Hierarchical High-Resolution Imaging
  Algorithm for mmWave FMCW Radar in Autonomous Driving}.
\newblock \bibinfo{journal}{\emph{IEEE Transactions on Vehicular Technology}}
  \bibinfo{volume}{70}, \bibinfo{number}{8} (\bibinfo{year}{2021}),
  \bibinfo{pages}{7322--7334}.
\newblock
\urldef\tempurl%
\url{https://doi.org/10.1109/TVT.2021.3092355}
\showDOI{\tempurl}


\bibitem[\protect\citeauthoryear{Goodin, Carruth, Doude, and Hudson}{Goodin
  et~al\mbox{.}}{2019}]%
        {goodin2019predicting}
\bibfield{author}{\bibinfo{person}{Christopher Goodin}, \bibinfo{person}{Daniel
  Carruth}, \bibinfo{person}{Matthew Doude}, {and} \bibinfo{person}{Christopher
  Hudson}.} \bibinfo{year}{2019}\natexlab{}.
\newblock \showarticletitle{Predicting the Influence of Rain on LIDAR in ADAS}.
\newblock \bibinfo{journal}{\emph{Electronics}} \bibinfo{volume}{8},
  \bibinfo{number}{1} (\bibinfo{year}{2019}), \bibinfo{pages}{89}.
\newblock
\urldef\tempurl%
\url{https://doi.org/10.3390/electronics8010089}
\showDOI{\tempurl}


\bibitem[\protect\citeauthoryear{Han, Chen, Hsiao, and Fu}{Han
  et~al\mbox{.}}{2020}]%
        {han2020using}
\bibfield{author}{\bibinfo{person}{Hsiang-Yu Han}, \bibinfo{person}{Yu-Chi
  Chen}, \bibinfo{person}{Pei-Yung Hsiao}, {and} \bibinfo{person}{Li-Chen Fu}.}
  \bibinfo{year}{2020}\natexlab{}.
\newblock \showarticletitle{Using channel-wise attention for deep CNN based
  real-time semantic segmentation with class-aware edge information}.
\newblock \bibinfo{journal}{\emph{IEEE Transactions on Intelligent
  Transportation Systems}} \bibinfo{volume}{22}, \bibinfo{number}{2}
  (\bibinfo{year}{2020}), \bibinfo{pages}{1041--1051}.
\newblock
\urldef\tempurl%
\url{https://doi.org/10.1109/TITS.2019.2962094}
\showDOI{\tempurl}


\bibitem[\protect\citeauthoryear{Hendrickx, Kim, Kozareva, Nakov, S{\'e}aghdha,
  Pad{\'o}, Pennacchiotti, Romano, and Szpakowicz}{Hendrickx
  et~al\mbox{.}}{2019}]%
        {hendrickx2019semeval}
\bibfield{author}{\bibinfo{person}{Iris Hendrickx}, \bibinfo{person}{Su~Nam
  Kim}, \bibinfo{person}{Zornitsa Kozareva}, \bibinfo{person}{Preslav Nakov},
  \bibinfo{person}{Diarmuid~O S{\'e}aghdha}, \bibinfo{person}{Sebastian
  Pad{\'o}}, \bibinfo{person}{Marco Pennacchiotti}, \bibinfo{person}{Lorenza
  Romano}, {and} \bibinfo{person}{Stan Szpakowicz}.}
  \bibinfo{year}{2019}\natexlab{}.
\newblock \showarticletitle{Semeval-2010 task 8: Multi-way classification of
  semantic relations between pairs of nominals}.
\newblock \bibinfo{journal}{\emph{arXiv preprint arXiv:1911.10422}}
  (\bibinfo{year}{2019}).
\newblock
\urldef\tempurl%
\url{https://doi.org/10.48550/arXiv.1911.10422}
\showDOI{\tempurl}


\bibitem[\protect\citeauthoryear{Huang, Zhou, Deng, and Li}{Huang
  et~al\mbox{.}}{2021b}]%
        {huang2021nighttime}
\bibfield{author}{\bibinfo{person}{Danyang Huang}, \bibinfo{person}{Zhiheng
  Zhou}, \bibinfo{person}{Ming Deng}, {and} \bibinfo{person}{Zhihao Li}.}
  \bibinfo{year}{2021}\natexlab{b}.
\newblock \showarticletitle{Nighttime vehicle detection based on direction
  attention network and bayes corner localization}.
\newblock \bibinfo{journal}{\emph{Journal of Intelligent \& Fuzzy Systems}}
  \bibinfo{number}{Preprint} (\bibinfo{year}{2021}), \bibinfo{pages}{1--19}.
\newblock
\urldef\tempurl%
\url{https://doi.org/10.3233/JIFS-202676}
\showDOI{\tempurl}


\bibitem[\protect\citeauthoryear{Huang, McGill, DeCastro, Fletcher, Leonard,
  Williams, and Rosman}{Huang et~al\mbox{.}}{2021a}]%
        {huang2021carpal}
\bibfield{author}{\bibinfo{person}{Xin Huang}, \bibinfo{person}{Stephen~G
  McGill}, \bibinfo{person}{Jonathan~A DeCastro}, \bibinfo{person}{Luke
  Fletcher}, \bibinfo{person}{John~J Leonard}, \bibinfo{person}{Brian~C
  Williams}, {and} \bibinfo{person}{Guy Rosman}.}
  \bibinfo{year}{2021}\natexlab{a}.
\newblock \showarticletitle{CARPAL: Confidence-aware intent recognition for
  parallel autonomy}.
\newblock \bibinfo{journal}{\emph{IEEE Robotics and Automation Letters}}
  \bibinfo{volume}{6}, \bibinfo{number}{3} (\bibinfo{year}{2021}),
  \bibinfo{pages}{4433--4440}.
\newblock
\urldef\tempurl%
\url{https://doi.org/10.1109/LRA.2021.3068894}
\showDOI{\tempurl}


\bibitem[\protect\citeauthoryear{Jin, Tefft, and Horrey}{Jin
  et~al\mbox{.}}{2019}]%
        {jin2019mining}
\bibfield{author}{\bibinfo{person}{Lydia Jin}, \bibinfo{person}{Brian~C Tefft},
  {and} \bibinfo{person}{William~J Horrey}.} \bibinfo{year}{2019}\natexlab{}.
\newblock \showarticletitle{Mining consumer complaints to identify unsuccessful
  interactions with advanced driver assistance systems}. In
  \bibinfo{booktitle}{\emph{Proceedings of the 11th International Conference on
  Automotive User Interfaces and Interactive Vehicular Applications: Adjunct
  Proceedings}}. \bibinfo{pages}{71--75}.
\newblock
\urldef\tempurl%
\url{https://doi.org/10.1145/3349263.3351313}
\showDOI{\tempurl}


\bibitem[\protect\citeauthoryear{Kang and Kum}{Kang and Kum}{2020}]%
        {kang2020camera}
\bibfield{author}{\bibinfo{person}{Daejun Kang} {and} \bibinfo{person}{Dongsuk
  Kum}.} \bibinfo{year}{2020}\natexlab{}.
\newblock \showarticletitle{Camera and radar sensor fusion for robust vehicle
  localization via vehicle part localization}.
\newblock \bibinfo{journal}{\emph{IEEE Access}}  \bibinfo{volume}{8}
  (\bibinfo{year}{2020}), \bibinfo{pages}{75223--75236}.
\newblock
\urldef\tempurl%
\url{https://doi.org/10.1109/ACCESS.2020.2985075}
\showDOI{\tempurl}


\bibitem[\protect\citeauthoryear{Khairdoost, Shirpour, Bauer, and
  Beauchemin}{Khairdoost et~al\mbox{.}}{2020}]%
        {khairdoost2020real}
\bibfield{author}{\bibinfo{person}{Nima Khairdoost}, \bibinfo{person}{Mohsen
  Shirpour}, \bibinfo{person}{Michael~A Bauer}, {and} \bibinfo{person}{Steven~S
  Beauchemin}.} \bibinfo{year}{2020}\natexlab{}.
\newblock \showarticletitle{Real-time driver maneuver prediction using LSTM}.
\newblock \bibinfo{journal}{\emph{IEEE Transactions on Intelligent Vehicles}}
  \bibinfo{volume}{5}, \bibinfo{number}{4} (\bibinfo{year}{2020}),
  \bibinfo{pages}{714--724}.
\newblock
\urldef\tempurl%
\url{https://doi.org/10.1109/TIV.2020.3003889}
\showDOI{\tempurl}


\bibitem[\protect\citeauthoryear{Khan, Ali, Hassan, Ali, Kazmi, and
  Zaheer}{Khan et~al\mbox{.}}{2020}]%
        {khan2020lane}
\bibfield{author}{\bibinfo{person}{Hussam~Ullah Khan},
  \bibinfo{person}{Afsheen~Rafaqat Ali}, \bibinfo{person}{Ali Hassan},
  \bibinfo{person}{Ahmed Ali}, \bibinfo{person}{Wajahat Kazmi}, {and}
  \bibinfo{person}{Aamer Zaheer}.} \bibinfo{year}{2020}\natexlab{}.
\newblock \showarticletitle{Lane detection using lane boundary marker network
  with road geometry constraints}. In \bibinfo{booktitle}{\emph{Proceedings of
  the IEEE/CVF Winter Conference on Applications of Computer Vision}}.
  \bibinfo{pages}{1834--1843}.
\newblock


\bibitem[\protect\citeauthoryear{Kidd, Cicchino, Reagan, and Kerfoot}{Kidd
  et~al\mbox{.}}{2017}]%
        {kidd2017driver}
\bibfield{author}{\bibinfo{person}{David~G Kidd}, \bibinfo{person}{Jessica~B
  Cicchino}, \bibinfo{person}{Ian~J Reagan}, {and} \bibinfo{person}{Laura~B
  Kerfoot}.} \bibinfo{year}{2017}\natexlab{}.
\newblock \showarticletitle{Driver trust in five driver assistance technologies
  following real-world use in four production vehicles}.
\newblock \bibinfo{journal}{\emph{Traffic injury prevention}}
  \bibinfo{volume}{18}, \bibinfo{number}{sup1} (\bibinfo{year}{2017}),
  \bibinfo{pages}{S44--S50}.
\newblock
\urldef\tempurl%
\url{https://doi.org/10.1080/15389588.2017.1297532}
\showDOI{\tempurl}


\bibitem[\protect\citeauthoryear{Kim, Kim, Yang, Kee, and Chung}{Kim
  et~al\mbox{.}}{2021}]%
        {kim2021lane}
\bibfield{author}{\bibinfo{person}{Dae~Jung Kim}, \bibinfo{person}{Jin~Sung
  Kim}, \bibinfo{person}{Jin~Ho Yang}, \bibinfo{person}{Seok~Cheol Kee}, {and}
  \bibinfo{person}{Chung~Choo Chung}.} \bibinfo{year}{2021}\natexlab{}.
\newblock \showarticletitle{Lane change intention classification of surrounding
  vehicles utilizing open set recognition}.
\newblock \bibinfo{journal}{\emph{IEEE Access}}  \bibinfo{volume}{9}
  (\bibinfo{year}{2021}), \bibinfo{pages}{57589--57602}.
\newblock
\urldef\tempurl%
\url{https://doi.org/10.1109/ACCESS.2021.3072413}
\showDOI{\tempurl}


\bibitem[\protect\citeauthoryear{Kim, Han, and Senouci}{Kim
  et~al\mbox{.}}{2018}]%
        {8436959}
\bibfield{author}{\bibinfo{person}{Jihun Kim}, \bibinfo{person}{Dong~Seog Han},
  {and} \bibinfo{person}{Benaoumeur Senouci}.} \bibinfo{year}{2018}\natexlab{}.
\newblock \showarticletitle{Radar and Vision Sensor Fusion for Object Detection
  in Autonomous Vehicle Surroundings}. In \bibinfo{booktitle}{\emph{2018 Tenth
  International Conference on Ubiquitous and Future Networks (ICUFN)}}.
  \bibinfo{pages}{76--78}.
\newblock
\urldef\tempurl%
\url{https://doi.org/10.1109/ICUFN.2018.8436959}
\showDOI{\tempurl}


\bibitem[\protect\citeauthoryear{Kotur, Luki{\'c}, Kruni{\'c}, and
  Luka{\v{c}}}{Kotur et~al\mbox{.}}{2021a}]%
        {kotur2021camera}
\bibfield{author}{\bibinfo{person}{Mila Kotur}, \bibinfo{person}{Nemanja
  Luki{\'c}}, \bibinfo{person}{Mom{\v{c}}ilo Kruni{\'c}}, {and}
  \bibinfo{person}{{\v{Z}}eljko Luka{\v{c}}}.}
  \bibinfo{year}{2021}\natexlab{a}.
\newblock \showarticletitle{Camera and LiDAR Sensor Fusion for 3D Object
  Tracking in a Collision Avoidance System}. In \bibinfo{booktitle}{\emph{2021
  Zooming Innovation in Consumer Technologies Conference (ZINC)}}. IEEE,
  \bibinfo{pages}{198--202}.
\newblock
\urldef\tempurl%
\url{https://doi.org/10.1109/ZINC52049.2021.9499281}
\showDOI{\tempurl}


\bibitem[\protect\citeauthoryear{Kotur, Lukić, Krunić, and Lukač}{Kotur
  et~al\mbox{.}}{2021b}]%
        {9499281}
\bibfield{author}{\bibinfo{person}{Mila Kotur}, \bibinfo{person}{Nemanja
  Lukić}, \bibinfo{person}{Momčilo Krunić}, {and} \bibinfo{person}{Željko
  Lukač}.} \bibinfo{year}{2021}\natexlab{b}.
\newblock \showarticletitle{Camera and LiDAR Sensor Fusion for 3D Object
  Tracking in a Collision Avoidance System}. In \bibinfo{booktitle}{\emph{2021
  Zooming Innovation in Consumer Technologies Conference (ZINC)}}.
  \bibinfo{pages}{198--202}.
\newblock
\urldef\tempurl%
\url{https://doi.org/10.1109/ZINC52049.2021.9499281}
\showDOI{\tempurl}


\bibitem[\protect\citeauthoryear{Kusano and Gabler}{Kusano and Gabler}{2012}]%
        {kusano2012safety}
\bibfield{author}{\bibinfo{person}{Kristofer~D Kusano} {and}
  \bibinfo{person}{Hampton~C Gabler}.} \bibinfo{year}{2012}\natexlab{}.
\newblock \showarticletitle{Safety benefits of forward collision warning, brake
  assist, and autonomous braking systems in rear-end collisions}.
\newblock \bibinfo{journal}{\emph{IEEE Transactions on Intelligent
  Transportation Systems}} \bibinfo{volume}{13}, \bibinfo{number}{4}
  (\bibinfo{year}{2012}), \bibinfo{pages}{1546--1555}.
\newblock
\urldef\tempurl%
\url{https://doi.org/10.1109/TITS.2012.2191542}
\showDOI{\tempurl}


\bibitem[\protect\citeauthoryear{Kusano and Gabler}{Kusano and Gabler}{2015}]%
        {kusano2015comparison}
\bibfield{author}{\bibinfo{person}{Kristofer~D Kusano} {and}
  \bibinfo{person}{Hampton~C Gabler}.} \bibinfo{year}{2015}\natexlab{}.
\newblock \showarticletitle{Comparison of expected crash and injury reduction
  from production forward collision and lane departure warning systems}.
\newblock \bibinfo{journal}{\emph{Traffic injury prevention}}
  \bibinfo{volume}{16}, \bibinfo{number}{sup2} (\bibinfo{year}{2015}),
  \bibinfo{pages}{S109--S114}.
\newblock
\urldef\tempurl%
\url{https://doi.org/10.1080/15389588.2015.1063619}
\showDOI{\tempurl}


\bibitem[\protect\citeauthoryear{Lee, Kim, Kim, and Lee}{Lee
  et~al\mbox{.}}{2021}]%
        {lee2021intelligent}
\bibfield{author}{\bibinfo{person}{Dongwook Lee}, \bibinfo{person}{Ji-Chul
  Kim}, \bibinfo{person}{Mingeuk Kim}, {and} \bibinfo{person}{Hanmin Lee}.}
  \bibinfo{year}{2021}\natexlab{}.
\newblock \showarticletitle{Intelligent tire sensor-based real-time road
  surface classification using an artificial neural network}.
\newblock \bibinfo{journal}{\emph{Sensors}} \bibinfo{volume}{21},
  \bibinfo{number}{9} (\bibinfo{year}{2021}), \bibinfo{pages}{3233}.
\newblock
\urldef\tempurl%
\url{https://doi.org/10.3390/s21093233}
\showDOI{\tempurl}


\bibitem[\protect\citeauthoryear{Li, Qu, Liu, Sun, and Wang}{Li
  et~al\mbox{.}}{2020}]%
        {li2020lane}
\bibfield{author}{\bibinfo{person}{Wenhui Li}, \bibinfo{person}{Feng Qu},
  \bibinfo{person}{Jialun Liu}, \bibinfo{person}{Fengdong Sun}, {and}
  \bibinfo{person}{Ying Wang}.} \bibinfo{year}{2020}\natexlab{}.
\newblock \showarticletitle{A lane detection network based on IBN and
  attention}.
\newblock \bibinfo{journal}{\emph{Multimedia Tools and Applications}}
  \bibinfo{volume}{79}, \bibinfo{number}{23} (\bibinfo{year}{2020}),
  \bibinfo{pages}{16473--16486}.
\newblock
\urldef\tempurl%
\url{https://doi.org/10.1007/s11042-019-7475-x}
\showDOI{\tempurl}


\bibitem[\protect\citeauthoryear{Lim, Islam, Kim, and Joung}{Lim
  et~al\mbox{.}}{2020}]%
        {9045356}
\bibfield{author}{\bibinfo{person}{Kiho Lim}, \bibinfo{person}{Tariqul Islam},
  \bibinfo{person}{Hyunbum Kim}, {and} \bibinfo{person}{Jingon Joung}.}
  \bibinfo{year}{2020}\natexlab{}.
\newblock \showarticletitle{A Sybil Attack Detection Scheme based on ADAS
  Sensors for Vehicular Networks}. In \bibinfo{booktitle}{\emph{2020 IEEE 17th
  Annual Consumer Communications \& Networking Conference (CCNC)}}.
  \bibinfo{pages}{1--5}.
\newblock
\urldef\tempurl%
\url{https://doi.org/10.1109/CCNC46108.2020.9045356}
\showDOI{\tempurl}


\bibitem[\protect\citeauthoryear{Lin, Dai, Wu, and Chen}{Lin
  et~al\mbox{.}}{2020}]%
        {lin2020vision}
\bibfield{author}{\bibinfo{person}{Huei-Yung Lin}, \bibinfo{person}{Jyun-Min
  Dai}, \bibinfo{person}{Lu-Ting Wu}, {and} \bibinfo{person}{Li-Qi Chen}.}
  \bibinfo{year}{2020}\natexlab{}.
\newblock \showarticletitle{A vision-based driver assistance system with
  forward collision and overtaking detection}.
\newblock \bibinfo{journal}{\emph{Sensors}} \bibinfo{volume}{20},
  \bibinfo{number}{18} (\bibinfo{year}{2020}), \bibinfo{pages}{5139}.
\newblock
\urldef\tempurl%
\url{https://doi.org/10.3390/s20185139}
\showDOI{\tempurl}


\bibitem[\protect\citeauthoryear{McDonald, Carney, and McGehee}{McDonald
  et~al\mbox{.}}{2018}]%
        {mcdonald2018vehicle}
\bibfield{author}{\bibinfo{person}{Ashley McDonald}, \bibinfo{person}{Cher
  Carney}, {and} \bibinfo{person}{Daniel~V McGehee}.}
  \bibinfo{year}{2018}\natexlab{}.
\newblock \showarticletitle{Vehicle owners' experiences with and reactions to
  advanced driver assistance systems}.
\newblock  (\bibinfo{year}{2018}).
\newblock


\bibitem[\protect\citeauthoryear{M{\"o}ller, Jehle, and Haas}{M{\"o}ller
  et~al\mbox{.}}{2018}]%
        {moller2018challenges}
\bibfield{author}{\bibinfo{person}{Dietmar~PF M{\"o}ller},
  \bibinfo{person}{Isabell~A Jehle}, {and} \bibinfo{person}{Roland~E Haas}.}
  \bibinfo{year}{2018}\natexlab{}.
\newblock \showarticletitle{Challenges for vehicular cybersecurity}. In
  \bibinfo{booktitle}{\emph{2018 IEEE International Conference on
  Electro/Information Technology (EIT)}}. IEEE, \bibinfo{pages}{0428--0433}.
\newblock
\urldef\tempurl%
\url{https://doi.org/10.1109/EIT.2018.8500208}
\showDOI{\tempurl}


\bibitem[\protect\citeauthoryear{Mueller, Reagan, and Cicchino}{Mueller
  et~al\mbox{.}}{2021}]%
        {mueller2021addressing}
\bibfield{author}{\bibinfo{person}{Alexandra~S Mueller}, \bibinfo{person}{Ian~J
  Reagan}, {and} \bibinfo{person}{Jessica~B Cicchino}.}
  \bibinfo{year}{2021}\natexlab{}.
\newblock \showarticletitle{Addressing Driver Disengagement and Proper System
  Use: Human Factors Recommendations for Level 2 Driving Automation Design}.
\newblock \bibinfo{journal}{\emph{Journal of Cognitive Engineering and Decision
  Making}} \bibinfo{volume}{15}, \bibinfo{number}{1} (\bibinfo{year}{2021}),
  \bibinfo{pages}{3--27}.
\newblock
\urldef\tempurl%
\url{https://doi.org/10.1177/1555343420983126}
\showDOI{\tempurl}


\bibitem[\protect\citeauthoryear{Nataprawira, Gu, Goncharenko, and
  Kamijo}{Nataprawira et~al\mbox{.}}{2021}]%
        {nataprawira2021pedestrian}
\bibfield{author}{\bibinfo{person}{Jason Nataprawira}, \bibinfo{person}{Yanlei
  Gu}, \bibinfo{person}{Igor Goncharenko}, {and} \bibinfo{person}{Shunsuke
  Kamijo}.} \bibinfo{year}{2021}\natexlab{}.
\newblock \showarticletitle{Pedestrian detection using multispectral images and
  a deep neural network}.
\newblock \bibinfo{journal}{\emph{Sensors}} \bibinfo{volume}{21},
  \bibinfo{number}{7} (\bibinfo{year}{2021}), \bibinfo{pages}{2536}.
\newblock
\urldef\tempurl%
\url{https://doi.org/10.3390/s21072536}
\showDOI{\tempurl}


\bibitem[\protect\citeauthoryear{NHTSA}{NHTSA}{2011}]%
        {NHTSA2011}
\bibfield{author}{\bibinfo{person}{NHTSA}.} \bibinfo{year}{2011}\natexlab{}.
\newblock \bibinfo{booktitle}{\emph{Vehicle Owner's Complaint Database}}.
\newblock
\urldef\tempurl%
\url{http://www-odi.nhtsa.dot.gov/downloads/}
\showURL{%
Retrieved October 7, 2011 from \tempurl}


\bibitem[\protect\citeauthoryear{Parasuraman and Riley}{Parasuraman and
  Riley}{1997}]%
        {parasuraman1997humans}
\bibfield{author}{\bibinfo{person}{Raja Parasuraman} {and}
  \bibinfo{person}{Victor Riley}.} \bibinfo{year}{1997}\natexlab{}.
\newblock \showarticletitle{Humans and automation: Use, misuse, disuse, abuse}.
\newblock \bibinfo{journal}{\emph{Human factors}} \bibinfo{volume}{39},
  \bibinfo{number}{2} (\bibinfo{year}{1997}), \bibinfo{pages}{230--253}.
\newblock
\urldef\tempurl%
\url{https://doi.org/10.1518/001872097778543886}
\showDOI{\tempurl}


\bibitem[\protect\citeauthoryear{Paul, Chauhan, Srivastava, and Baruah}{Paul
  et~al\mbox{.}}{[n.d.]}]%
        {pauladvanced}
\bibfield{author}{\bibinfo{person}{A Paul}, \bibinfo{person}{R Chauhan},
  \bibinfo{person}{R Srivastava}, {and} \bibinfo{person}{M Baruah}.}
  \bibinfo{year}{[n.d.]}\natexlab{}.
\newblock \showarticletitle{Advanced driver assistance systems. 2016}.
\newblock \bibinfo{journal}{\emph{SAE Technical Paper}}
  (\bibinfo{year}{[n.\,d.]}).
\newblock


\bibitem[\protect\citeauthoryear{Piccoli, Balakrishnan, Perez, Sachdeo, Nunez,
  Tang, Andreasson, Bjurek, Raj, Davidsson, et~al\mbox{.}}{Piccoli
  et~al\mbox{.}}{2020}]%
        {piccoli2020fussi}
\bibfield{author}{\bibinfo{person}{Francesco Piccoli},
  \bibinfo{person}{Rajarathnam Balakrishnan}, \bibinfo{person}{Maria~Jesus
  Perez}, \bibinfo{person}{Moraldeepsingh Sachdeo}, \bibinfo{person}{Carlos
  Nunez}, \bibinfo{person}{Matthew Tang}, \bibinfo{person}{Kajsa Andreasson},
  \bibinfo{person}{Kalle Bjurek}, \bibinfo{person}{Ria~Dass Raj},
  \bibinfo{person}{Ebba Davidsson}, {et~al\mbox{.}}}
  \bibinfo{year}{2020}\natexlab{}.
\newblock \showarticletitle{Fussi-net: Fusion of spatio-temporal skeletons for
  intention prediction network}. In \bibinfo{booktitle}{\emph{2020 54th
  Asilomar Conference on Signals, Systems, and Computers}}. IEEE,
  \bibinfo{pages}{68--72}.
\newblock
\urldef\tempurl%
\url{https://doi.org/10.1109/IEEECONF51394.2020.9443552}
\showDOI{\tempurl}


\bibitem[\protect\citeauthoryear{Reagan and McCartt}{Reagan and
  McCartt}{2016}]%
        {reagan2016observed}
\bibfield{author}{\bibinfo{person}{Ian~J Reagan} {and} \bibinfo{person}{Anne~T
  McCartt}.} \bibinfo{year}{2016}\natexlab{}.
\newblock \showarticletitle{Observed activation status of lane departure
  warning and forward collision warning of Honda vehicles at dealership service
  centers}.
\newblock \bibinfo{journal}{\emph{Traffic injury prevention}}
  \bibinfo{volume}{17}, \bibinfo{number}{8} (\bibinfo{year}{2016}),
  \bibinfo{pages}{827--832}.
\newblock
\urldef\tempurl%
\url{https://doi.org/10.1080/15389588.2016.1149698}
\showDOI{\tempurl}


\bibitem[\protect\citeauthoryear{Richardson}{Richardson}{2001}]%
        {richardson2001human}
\bibfield{author}{\bibinfo{person}{JH Richardson}.}
  \bibinfo{year}{2001}\natexlab{}.
\newblock \showarticletitle{Human factors research priorities for ADAS systems:
  a UK perspective}. In \bibinfo{booktitle}{\emph{2001 ADAS. International
  Conference on Advanced Driver Assistance Systems,(IEE Conf. Publ. No. 483)}}.
  IET, \bibinfo{pages}{20--24}.
\newblock


\bibitem[\protect\citeauthoryear{Roh, Kim, and Im}{Roh et~al\mbox{.}}{2020}]%
        {roh2020analysis}
\bibfield{author}{\bibinfo{person}{Chang-Gyun Roh}, \bibinfo{person}{Jisoo
  Kim}, {and} \bibinfo{person}{I-Jeong Im}.} \bibinfo{year}{2020}\natexlab{}.
\newblock \showarticletitle{Analysis of impact of rain conditions on ADAS}.
\newblock \bibinfo{journal}{\emph{Sensors}} \bibinfo{volume}{20},
  \bibinfo{number}{23} (\bibinfo{year}{2020}), \bibinfo{pages}{6720}.
\newblock
\urldef\tempurl%
\url{https://doi.org/10.3390/s20236720}
\showDOI{\tempurl}


\bibitem[\protect\citeauthoryear{{\v{S}}abanovi{\v{c}}, {\v{Z}}uraulis,
  Prentkovskis, and Skrickij}{{\v{S}}abanovi{\v{c}} et~al\mbox{.}}{2020}]%
        {vsabanovivc2020identification}
\bibfield{author}{\bibinfo{person}{Eldar {\v{S}}abanovi{\v{c}}},
  \bibinfo{person}{Vidas {\v{Z}}uraulis}, \bibinfo{person}{Olegas
  Prentkovskis}, {and} \bibinfo{person}{Viktor Skrickij}.}
  \bibinfo{year}{2020}\natexlab{}.
\newblock \showarticletitle{Identification of road-surface type using deep
  neural networks for friction coefficient estimation}.
\newblock \bibinfo{journal}{\emph{Sensors}} \bibinfo{volume}{20},
  \bibinfo{number}{3} (\bibinfo{year}{2020}), \bibinfo{pages}{612}.
\newblock
\urldef\tempurl%
\url{https://doi.org/10.3390/s20030612}
\showDOI{\tempurl}


\bibitem[\protect\citeauthoryear{Seo, Oh, and Kim}{Seo et~al\mbox{.}}{2021}]%
        {seo2021study}
\bibfield{author}{\bibinfo{person}{J Seo}, \bibinfo{person}{S Oh}, {and}
  \bibinfo{person}{YK Kim}.} \bibinfo{year}{2021}\natexlab{}.
\newblock \showarticletitle{A study of curved lane detection based on dual
  sensor monitoring of LiDAR and camera}.
\newblock \bibinfo{journal}{\emph{Transaction of the Korean Society of
  Automotive Engineers}} \bibinfo{volume}{29}, \bibinfo{number}{2}
  (\bibinfo{year}{2021}), \bibinfo{pages}{197--204}.
\newblock


\bibitem[\protect\citeauthoryear{Song, Tan, Qin, Lu, and Liu}{Song
  et~al\mbox{.}}{2020}]%
        {song2020mpnet}
\bibfield{author}{\bibinfo{person}{Kaitao Song}, \bibinfo{person}{Xu Tan},
  \bibinfo{person}{Tao Qin}, \bibinfo{person}{Jianfeng Lu}, {and}
  \bibinfo{person}{Tie-Yan Liu}.} \bibinfo{year}{2020}\natexlab{}.
\newblock \showarticletitle{Mpnet: Masked and permuted pre-training for
  language understanding}.
\newblock \bibinfo{journal}{\emph{Advances in Neural Information Processing
  Systems}}  \bibinfo{volume}{33} (\bibinfo{year}{2020}),
  \bibinfo{pages}{16857--16867}.
\newblock


\bibitem[\protect\citeauthoryear{Stanton and Young}{Stanton and Young}{2005}]%
        {stanton2005driver}
\bibfield{author}{\bibinfo{person}{Neville~A Stanton} {and}
  \bibinfo{person}{Mark~S Young}.} \bibinfo{year}{2005}\natexlab{}.
\newblock \showarticletitle{Driver behaviour with adaptive cruise control}.
\newblock \bibinfo{journal}{\emph{Ergonomics}} \bibinfo{volume}{48},
  \bibinfo{number}{10} (\bibinfo{year}{2005}), \bibinfo{pages}{1294--1313}.
\newblock
\urldef\tempurl%
\url{https://doi.org/10.1080/00140130500252990}
\showDOI{\tempurl}


\bibitem[\protect\citeauthoryear{Sun, Nakane, Zhang, and Zhang}{Sun
  et~al\mbox{.}}{2021}]%
        {sun2021evolutionary}
\bibfield{author}{\bibinfo{person}{Haitian Sun}, \bibinfo{person}{Takumi
  Nakane}, \bibinfo{person}{Naidan Zhang}, {and} \bibinfo{person}{Chao Zhang}.}
  \bibinfo{year}{2021}\natexlab{}.
\newblock \showarticletitle{Evolutionary Rear-Lamp Tracking at Nighttime}.
\newblock \bibinfo{journal}{\emph{IEEE Access}}  \bibinfo{volume}{9}
  (\bibinfo{year}{2021}), \bibinfo{pages}{86667--86676}.
\newblock
\urldef\tempurl%
\url{https://doi.org/10.1109/ACCESS.2021.3087238}
\showDOI{\tempurl}


\bibitem[\protect\citeauthoryear{Teo, Sutopo, Lim, and Wong}{Teo
  et~al\mbox{.}}{2021}]%
        {teo2021innovative}
\bibfield{author}{\bibinfo{person}{Ting~Yau Teo}, \bibinfo{person}{Ricky
  Sutopo}, \bibinfo{person}{Joanne Mun-Yee Lim}, {and}
  \bibinfo{person}{KokSheik Wong}.} \bibinfo{year}{2021}\natexlab{}.
\newblock \showarticletitle{Innovative lane detection method to increase the
  accuracy of lane departure warning system}.
\newblock \bibinfo{journal}{\emph{Multimedia Tools and Applications}}
  \bibinfo{volume}{80}, \bibinfo{number}{2} (\bibinfo{year}{2021}),
  \bibinfo{pages}{2063--2080}.
\newblock
\urldef\tempurl%
\url{https://doi.org/10.1007/s11042-020-09819-0}
\showDOI{\tempurl}


\bibitem[\protect\citeauthoryear{Teoh}{Teoh}{2021}]%
        {teoh2021effectiveness}
\bibfield{author}{\bibinfo{person}{Eric~R Teoh}.}
  \bibinfo{year}{2021}\natexlab{}.
\newblock \showarticletitle{Effectiveness of front crash prevention systems in
  reducing large truck real-world crash rates}.
\newblock \bibinfo{journal}{\emph{Traffic injury prevention}}
  \bibinfo{volume}{22}, \bibinfo{number}{4} (\bibinfo{year}{2021}),
  \bibinfo{pages}{284--289}.
\newblock
\urldef\tempurl%
\url{https://doi.org/10.1080/15389588.2021.1893700}
\showDOI{\tempurl}


\bibitem[\protect\citeauthoryear{Tian, Liu, Zhong, and Zeng}{Tian
  et~al\mbox{.}}{2021}]%
        {tian2021lsd}
\bibfield{author}{\bibinfo{person}{Jun Tian}, \bibinfo{person}{Shiwang Liu},
  \bibinfo{person}{Xunyu Zhong}, {and} \bibinfo{person}{Jianping Zeng}.}
  \bibinfo{year}{2021}\natexlab{}.
\newblock \showarticletitle{LSD-based adaptive lane detection and tracking for
  ADAS in structured road environment}.
\newblock \bibinfo{journal}{\emph{Soft Computing}} \bibinfo{volume}{25},
  \bibinfo{number}{7} (\bibinfo{year}{2021}), \bibinfo{pages}{5709--5722}.
\newblock
\urldef\tempurl%
\url{https://doi.org/10.1007/s00500-020-05566-4}
\showDOI{\tempurl}


\bibitem[\protect\citeauthoryear{Tumas, Serackis, and Nowosielski}{Tumas
  et~al\mbox{.}}{2021}]%
        {tumas2021augmentation}
\bibfield{author}{\bibinfo{person}{Paulius Tumas}, \bibinfo{person}{Art{\=u}ras
  Serackis}, {and} \bibinfo{person}{Adam Nowosielski}.}
  \bibinfo{year}{2021}\natexlab{}.
\newblock \showarticletitle{Augmentation of Severe Weather Impact to
  Far-Infrared Sensor Images to Improve Pedestrian Detection System}.
\newblock \bibinfo{journal}{\emph{Electronics}} \bibinfo{volume}{10},
  \bibinfo{number}{8} (\bibinfo{year}{2021}), \bibinfo{pages}{934}.
\newblock
\urldef\tempurl%
\url{https://doi.org/10.3390/electronics10080934}
\showDOI{\tempurl}


\bibitem[\protect\citeauthoryear{Umehira, Okuda, Wang, Takeda, and
  Kuroda}{Umehira et~al\mbox{.}}{2020}]%
        {9266712}
\bibfield{author}{\bibinfo{person}{Masahiro Umehira}, \bibinfo{person}{Takeo
  Okuda}, \bibinfo{person}{Xiaoyan Wang}, \bibinfo{person}{Shigeki Takeda},
  {and} \bibinfo{person}{Hiroshi Kuroda}.} \bibinfo{year}{2020}\natexlab{}.
\newblock \showarticletitle{An Adaptive Interference Detection and Suppression
  Scheme Using Iterative Processing for Automotive FMCW Radars}. In
  \bibinfo{booktitle}{\emph{2020 IEEE Radar Conference (RadarConf20)}}.
  \bibinfo{pages}{1--5}.
\newblock
\urldef\tempurl%
\url{https://doi.org/10.1109/RadarConf2043947.2020.9266712}
\showDOI{\tempurl}


\bibitem[\protect\citeauthoryear{Vermiglio, Champaney, Sancarlos, Daim, Kedzia,
  Duval, Diez, and Chinesta}{Vermiglio et~al\mbox{.}}{2020}]%
        {2020}
\bibfield{author}{\bibinfo{person}{Simona Vermiglio}, \bibinfo{person}{Victor
  Champaney}, \bibinfo{person}{Abel Sancarlos}, \bibinfo{person}{Fatima Daim},
  \bibinfo{person}{Jean~Claude Kedzia}, \bibinfo{person}{Jean~Louis Duval},
  \bibinfo{person}{Pedro Diez}, {and} \bibinfo{person}{Francisco Chinesta}.}
  \bibinfo{year}{2020}\natexlab{}.
\newblock \showarticletitle{Parametric Electromagnetic Analysis of Radar-Based
  Advanced Driver Assistant Systems}.
\newblock \bibinfo{journal}{\emph{Sensors}} \bibinfo{volume}{20},
  \bibinfo{number}{19} (\bibinfo{date}{Oct} \bibinfo{year}{2020}),
  \bibinfo{pages}{5686}.
\newblock
\showISSN{1424-8220}
\urldef\tempurl%
\url{https://doi.org/10.3390/s20195686}
\showDOI{\tempurl}


\bibitem[\protect\citeauthoryear{Wang, Fu, Lai, Xu, Shi, and Wang}{Wang
  et~al\mbox{.}}{2018}]%
        {wang2018efficient}
\bibfield{author}{\bibinfo{person}{Yao Wang}, \bibinfo{person}{Fangfa Fu},
  \bibinfo{person}{Fengchang Lai}, \bibinfo{person}{Weizhe Xu},
  \bibinfo{person}{Jinjin Shi}, {and} \bibinfo{person}{Jinxiang Wang}.}
  \bibinfo{year}{2018}\natexlab{}.
\newblock \showarticletitle{Efficient road specular reflection removal based on
  gradient properties}.
\newblock \bibinfo{journal}{\emph{Multimedia Tools and Applications}}
  \bibinfo{volume}{77}, \bibinfo{number}{23} (\bibinfo{year}{2018}),
  \bibinfo{pages}{30615--30631}.
\newblock
\urldef\tempurl%
\url{https://doi.org/10.1007/s11042-018-6156-5}
\showDOI{\tempurl}


\bibitem[\protect\citeauthoryear{Wang, Liao, Wang, Oswald, Wu, Boriboonsomsin,
  Barth, Han, Kim, and Tiwari}{Wang et~al\mbox{.}}{2020}]%
        {wang2020driver}
\bibfield{author}{\bibinfo{person}{Ziran Wang}, \bibinfo{person}{Xishun Liao},
  \bibinfo{person}{Chao Wang}, \bibinfo{person}{David Oswald},
  \bibinfo{person}{Guoyuan Wu}, \bibinfo{person}{Kanok Boriboonsomsin},
  \bibinfo{person}{Matthew~J Barth}, \bibinfo{person}{Kyungtae Han},
  \bibinfo{person}{BaekGyu Kim}, {and} \bibinfo{person}{Prashant Tiwari}.}
  \bibinfo{year}{2020}\natexlab{}.
\newblock \showarticletitle{Driver behavior modeling using game engine and real
  vehicle: A learning-based approach}.
\newblock \bibinfo{journal}{\emph{IEEE Transactions on Intelligent Vehicles}}
  \bibinfo{volume}{5}, \bibinfo{number}{4} (\bibinfo{year}{2020}),
  \bibinfo{pages}{738--749}.
\newblock
\urldef\tempurl%
\url{https://doi.org/10.1109/TIV.2020.2991948}
\showDOI{\tempurl}


\bibitem[\protect\citeauthoryear{Wielitzka, Dagen, and Ortmaier}{Wielitzka
  et~al\mbox{.}}{2017}]%
        {wielitzka2017state}
\bibfield{author}{\bibinfo{person}{Mark Wielitzka}, \bibinfo{person}{Matthias
  Dagen}, {and} \bibinfo{person}{Tobias Ortmaier}.}
  \bibinfo{year}{2017}\natexlab{}.
\newblock \showarticletitle{State and maximum friction coefficient estimation
  in vehicle dynamics using UKF}. In \bibinfo{booktitle}{\emph{2017 American
  Control Conference (ACC)}}. IEEE, \bibinfo{pages}{4322--4327}.
\newblock
\urldef\tempurl%
\url{https://doi.org/10.23919/ACC.2017.7963620}
\showDOI{\tempurl}


\bibitem[\protect\citeauthoryear{Winner, Hakuli, Lotz, and Singer}{Winner
  et~al\mbox{.}}{2014}]%
        {winner2014handbook}
\bibfield{author}{\bibinfo{person}{Hermann Winner}, \bibinfo{person}{Stephan
  Hakuli}, \bibinfo{person}{Felix Lotz}, {and} \bibinfo{person}{Christina
  Singer}.} \bibinfo{year}{2014}\natexlab{}.
\newblock \bibinfo{booktitle}{\emph{Handbook of driver assistance systems}}.
\newblock \bibinfo{publisher}{Springer International Publishing Amsterdam, The
  Netherlands:}.
\newblock


\bibitem[\protect\citeauthoryear{Yuan, Wang, Peng, and Fu}{Yuan
  et~al\mbox{.}}{2021}]%
        {yuan2021novel}
\bibfield{author}{\bibinfo{person}{Guoliang Yuan}, \bibinfo{person}{Yafei
  Wang}, \bibinfo{person}{Jinjia Peng}, {and} \bibinfo{person}{Xianping Fu}.}
  \bibinfo{year}{2021}\natexlab{}.
\newblock \showarticletitle{A Novel Driving Behavior Learning and Visualization
  Method With Natural Gaze Prediction}.
\newblock \bibinfo{journal}{\emph{IEEE Access}}  \bibinfo{volume}{9}
  (\bibinfo{year}{2021}), \bibinfo{pages}{18560--18568}.
\newblock
\urldef\tempurl%
\url{https://doi.org/10.1109/ACCESS.2021.3054951}
\showDOI{\tempurl}


\bibitem[\protect\citeauthoryear{Zhang, Wu, Gou, and Chen}{Zhang
  et~al\mbox{.}}{2021a}]%
        {zhang2021rs}
\bibfield{author}{\bibinfo{person}{Ronghui Zhang}, \bibinfo{person}{Yueying
  Wu}, \bibinfo{person}{Wanting Gou}, {and} \bibinfo{person}{Junzhou Chen}.}
  \bibinfo{year}{2021}\natexlab{a}.
\newblock \showarticletitle{Rs-Lane: a robust lane detection method based on
  ResNeSt and self-attention distillation for challenging traffic situations}.
\newblock \bibinfo{journal}{\emph{Journal of advanced transportation}}
  \bibinfo{volume}{2021} (\bibinfo{year}{2021}).
\newblock
\urldef\tempurl%
\url{https://doi.org/10.1155/2021/7544355}
\showDOI{\tempurl}


\bibitem[\protect\citeauthoryear{Zhang, Yang, and Zhou}{Zhang
  et~al\mbox{.}}{2021b}]%
        {zhang2021disengagement}
\bibfield{author}{\bibinfo{person}{Yangtao Zhang}, \bibinfo{person}{X~Jessie
  Yang}, {and} \bibinfo{person}{Feng Zhou}.} \bibinfo{year}{2021}\natexlab{b}.
\newblock \showarticletitle{Disengagement Cause-and-Effect Relationships
  Extraction Using an NLP Pipeline}.
\newblock \bibinfo{journal}{\emph{arXiv preprint arXiv:2111.03511}}
  (\bibinfo{year}{2021}).
\newblock
\urldef\tempurl%
\url{https://doi.org/10.48550/arXiv.2111.03511}
\showDOI{\tempurl}


\bibitem[\protect\citeauthoryear{Zhao, He, Yang, and Tao}{Zhao
  et~al\mbox{.}}{2022}]%
        {zhao2022research}
\bibfield{author}{\bibinfo{person}{Guangzhe Zhao}, \bibinfo{person}{Yanqing
  He}, \bibinfo{person}{Hanting Yang}, {and} \bibinfo{person}{Yong Tao}.}
  \bibinfo{year}{2022}\natexlab{}.
\newblock \showarticletitle{Research on fatigue detection based on visual
  features}.
\newblock \bibinfo{journal}{\emph{IET Image Processing}} \bibinfo{volume}{16},
  \bibinfo{number}{4} (\bibinfo{year}{2022}), \bibinfo{pages}{1044--1053}.
\newblock
\urldef\tempurl%
\url{https://doi.org/10.1049/ipr2.12207}
\showDOI{\tempurl}


\bibitem[\protect\citeauthoryear{Zheng and Zhao}{Zheng and Zhao}{2020}]%
        {zheng2020investigation}
\bibfield{author}{\bibinfo{person}{Hongyu Zheng} {and} \bibinfo{person}{Mingxin
  Zhao}.} \bibinfo{year}{2020}\natexlab{}.
\newblock \showarticletitle{An investigation on coordination of lane departure
  warning based on driver behaviour characteristics}.
\newblock \bibinfo{journal}{\emph{International Journal of Vehicle Autonomous
  Systems}} \bibinfo{volume}{15}, \bibinfo{number}{1} (\bibinfo{year}{2020}),
  \bibinfo{pages}{77--99}.
\newblock


\bibitem[\protect\citeauthoryear{Zhou, Liu, Ma, Wang, Zhang, and Dong}{Zhou
  et~al\mbox{.}}{2020b}]%
        {zhou2020driving}
\bibfield{author}{\bibinfo{person}{Dong Zhou}, \bibinfo{person}{Hongyi Liu},
  \bibinfo{person}{Huimin Ma}, \bibinfo{person}{Xiang Wang},
  \bibinfo{person}{Xiaoqin Zhang}, {and} \bibinfo{person}{Yuhan Dong}.}
  \bibinfo{year}{2020}\natexlab{b}.
\newblock \showarticletitle{Driving behavior prediction considering cognitive
  prior and driving context}.
\newblock \bibinfo{journal}{\emph{IEEE Transactions on Intelligent
  Transportation Systems}} \bibinfo{volume}{22}, \bibinfo{number}{5}
  (\bibinfo{year}{2020}), \bibinfo{pages}{2669--2678}.
\newblock
\urldef\tempurl%
\url{https://doi.org/10.1109/TITS.2020.2973751}
\showDOI{\tempurl}


\bibitem[\protect\citeauthoryear{Zhou, Alsaid, Blommer, Curry, Swaminathan,
  Kochhar, Talamonti, and Tijerina}{Zhou et~al\mbox{.}}{2022}]%
        {zhoupredicting2022}
\bibfield{author}{\bibinfo{person}{Feng Zhou}, \bibinfo{person}{Areen Alsaid},
  \bibinfo{person}{Mike Blommer}, \bibinfo{person}{Reates Curry},
  \bibinfo{person}{Radhakrishnan Swaminathan}, \bibinfo{person}{Dev Kochhar},
  \bibinfo{person}{Walter Talamonti}, {and} \bibinfo{person}{Louis Tijerina}.}
  \bibinfo{year}{2022}\natexlab{}.
\newblock \showarticletitle{Predicting Driver Fatigue in Monotonous Automated
  Driving with Explanation using GPBoost and SHAP}.
\newblock \bibinfo{journal}{\emph{International Journal of Human–Computer
  Interaction}} \bibinfo{volume}{38}, \bibinfo{number}{8}
  (\bibinfo{year}{2022}), \bibinfo{pages}{719--729}.
\newblock
\urldef\tempurl%
\url{https://doi.org/10.1080/10447318.2021.1965774}
\showDOI{\tempurl}


\bibitem[\protect\citeauthoryear{Zhou, Alsaid, Blommer, Curry, Swaminathan,
  Kochhar, Talamonti, Tijerina, and Lei}{Zhou et~al\mbox{.}}{2020a}]%
        {zhou2020driver}
\bibfield{author}{\bibinfo{person}{Feng Zhou}, \bibinfo{person}{Areen Alsaid},
  \bibinfo{person}{Mike Blommer}, \bibinfo{person}{Reates Curry},
  \bibinfo{person}{Radhakrishnan Swaminathan}, \bibinfo{person}{Dev Kochhar},
  \bibinfo{person}{Walter Talamonti}, \bibinfo{person}{Louis Tijerina}, {and}
  \bibinfo{person}{Baiying Lei}.} \bibinfo{year}{2020}\natexlab{a}.
\newblock \showarticletitle{Driver fatigue transition prediction in highly
  automated driving using physiological features}.
\newblock \bibinfo{journal}{\emph{Expert Systems with Applications}}
  \bibinfo{volume}{147} (\bibinfo{year}{2020}), \bibinfo{pages}{113204}.
\newblock
\urldef\tempurl%
\url{https://doi.org/10.1016/j.eswa.2020.113204}
\showDOI{\tempurl}


\bibitem[\protect\citeauthoryear{Zhou, Qu, Helander, and Jiao}{Zhou
  et~al\mbox{.}}{2011}]%
        {zhou2011affect}
\bibfield{author}{\bibinfo{person}{Feng Zhou}, \bibinfo{person}{Xingda Qu},
  \bibinfo{person}{Martin~G Helander}, {and} \bibinfo{person}{Jianxin~Roger
  Jiao}.} \bibinfo{year}{2011}\natexlab{}.
\newblock \showarticletitle{Affect prediction from physiological measures via
  visual stimuli}.
\newblock \bibinfo{journal}{\emph{International Journal of Human-Computer
  Studies}} \bibinfo{volume}{69}, \bibinfo{number}{12} (\bibinfo{year}{2011}),
  \bibinfo{pages}{801--819}.
\newblock
\urldef\tempurl%
\url{https://doi.org/10.1016/j.ijhcs.2011.07.005}
\showDOI{\tempurl}


\bibitem[\protect\citeauthoryear{Zhou, Yang, and Zhang}{Zhou
  et~al\mbox{.}}{2020c}]%
        {zhou2020takeover}
\bibfield{author}{\bibinfo{person}{Feng Zhou}, \bibinfo{person}{X~Jessie Yang},
  {and} \bibinfo{person}{Xin Zhang}.} \bibinfo{year}{2020}\natexlab{c}.
\newblock \showarticletitle{Takeover transition in autonomous vehicles: a
  YouTube study}.
\newblock \bibinfo{journal}{\emph{International Journal of Human--Computer
  Interaction}} \bibinfo{volume}{36}, \bibinfo{number}{3}
  (\bibinfo{year}{2020}), \bibinfo{pages}{295--306}.
\newblock
\urldef\tempurl%
\url{https://doi.org/10.1080/10447318.2019.1634317}
\showDOI{\tempurl}


\bibitem[\protect\citeauthoryear{Zhu, Ferrante, Karlsson, and Zorzi}{Zhu
  et~al\mbox{.}}{2019}]%
        {zhu2019fusion}
\bibfield{author}{\bibinfo{person}{Bin Zhu}, \bibinfo{person}{Augusto
  Ferrante}, \bibinfo{person}{Johan Karlsson}, {and} \bibinfo{person}{Mattia
  Zorzi}.} \bibinfo{year}{2019}\natexlab{}.
\newblock \showarticletitle{Fusion of sensors data in automotive radar systems:
  A spectral estimation approach}. In \bibinfo{booktitle}{\emph{2019 IEEE 58th
  Conference on Decision and Control (CDC)}}. IEEE,
  \bibinfo{pages}{5088--5093}.
\newblock
\urldef\tempurl%
\url{https://doi.org/10.1109/CDC40024.2019.9029655}
\showDOI{\tempurl}


\bibitem[\protect\citeauthoryear{Ziebinski, Cupek, Erdogan, and
  Waechter}{Ziebinski et~al\mbox{.}}{2016}]%
        {ziebinski2016survey}
\bibfield{author}{\bibinfo{person}{Adam Ziebinski}, \bibinfo{person}{Rafal
  Cupek}, \bibinfo{person}{Hueseyin Erdogan}, {and} \bibinfo{person}{Sonja
  Waechter}.} \bibinfo{year}{2016}\natexlab{}.
\newblock \showarticletitle{A survey of ADAS technologies for the future
  perspective of sensor fusion}. In \bibinfo{booktitle}{\emph{International
  Conference on Computational Collective Intelligence}}. Springer,
  \bibinfo{pages}{135--146}.
\newblock
\urldef\tempurl%
\url{https://doi.org/10.1007/978-3-319-45246-3_13}
\showDOI{\tempurl}


\end{thebibliography}

\end{document}